\documentclass[runningheads]{llncs}
\usepackage{graphicx}
\usepackage{tikz}
\usepackage{comment}
\usepackage{amsmath,amssymb} % define this before the line numbering.
\usepackage[accsupp]{axessibility}  % Improves PDF readability for those with disabilities.

\usepackage[width=122mm,left=12mm,paperwidth=146mm,height=193mm,top=12mm,paperheight=217mm]{geometry}

\usepackage{algorithm}
\usepackage{algpseudocode}% http://ctan.org/pkg/algorithmicx

\newcommand{\cmark}{\ding{51}}%
\newcommand{\xmark}{\ding{55}}%
\usepackage{multirow}
\usepackage{stmaryrd}
\usepackage{tabularx}
\usepackage{stmaryrd}
\usepackage{trimclip}
\usepackage{xspace}
\makeatletter
\DeclareRobustCommand\onedot{\futurelet\@let@token\@onedot}
\def\@onedot{\ifx\@let@token.\else.\null\fi\xspace}
\def\eg{\emph{e.g}\onedot} 
\def\ie{\emph{i.e}\onedot} 
 
\def\etc{\emph{etc}\onedot} 
\def\wrt{w.r.t\onedot} 
\def\etal{\emph{et al}\onedot}
\DeclareRobustCommand{\shortto}{%
  \mathrel{\mathpalette\short@to\relax}%
}
\newcommand{\short@to}[2]{%
  \mkern2mu
  \clipbox{{.5\width} 0 0 0}{$\m@th#1\vphantom{+}{\shortrightarrow}$}%
  }

\usepackage{wrapfig}
\usepackage{makecell}
\usepackage{booktabs}
\usepackage{xcolor}
\usepackage{color, colortbl}
\usepackage{pifont}% http://ctan.org/pkg/pifont

\usepackage{gensymb}

\definecolor{lightgreen}{RGB}{100,220,100}
\definecolor{darkgreen}{RGB}{30,150,30}
\definecolor{darkblue}{RGB}{0,0,127}
\definecolor{darkyellow}{RGB}{171,133,0}
\definecolor{darkred}{RGB}{180,20,20}
\definecolor{darkmagenta}{RGB}{127,0,127}
\definecolor{darkcyan}{RGB}{0,127,127}

\definecolor{purple}{HTML}{9900ff}
\definecolor{darkpink}{HTML}{ff00ff}
\definecolor{maroon}{HTML}{980000}

\newcommand{\ak}[1]{\textcolor{red}{#1}} % Akshay
\definecolor{coolblack}{rgb}{0.0, 0.23, 0.64}
\newcommand{\jnkc}[1]{\textcolor{coolblack}{#1}} % Jogendra

\usepackage{lipsum}
\usepackage{bbm}

\newcommand{\expectation}{\mathop{\mathbb{E}}}

\definecolor{darkpastelgreen}{rgb}{0.01, 0.75, 0.24}

\usepackage[pagebackref=true,breaklinks=true,colorlinks,citecolor=darkpastelgreen,bookmarks=false]{hyperref}

\newcommand\blfootnote[1]{%
  \begingroup
  \renewcommand\thefootnote{}\footnote{#1}%
  \addtocounter{footnote}{-1}%
  \endgroup
}

\begin{document}
% \linenumbers
\pagestyle{headings}
\mainmatter
\def\ECCVSubNumber{912}  % Insert your submission number here
% Concurrent subsidiary supervision for unsupervised source-free domain adaptation
% Unsupervised Source-Free Domain Adaptation with Subsidiary Supervision

\title{Concurrent Subsidiary Supervision for Unsupervised Source-Free Domain Adaptation} % Replace with your title

\begin{comment}
\titlerunning{ECCV-22 submission ID \ECCVSubNumber} 
\authorrunning{ECCV-22 submission ID \ECCVSubNumber} 
\author{Anonymous ECCV submission}
\institute{Paper ID \ECCVSubNumber}
\end{comment}
%******************

% \begin{comment}
\titlerunning{Concurrent Subsidiary Supervision for Unsupervised Source-Free DA}
\author{Jogendra Nath Kundu\inst{1}* \and
Suvaansh Bhambri\inst{1}* \and
Akshay Kulkarni\inst{1}*
\and \\
Hiran Sarkar\inst{1}
\and
Varun Jampani\inst{2}
\and
R. Venkatesh Babu\inst{1}
}
\authorrunning{J. N. Kundu et al.}
\institute{Indian Institute of Science \and Google Research}
% \end{comment}
%******************
\maketitle

\begin{abstract}
The prime challenge in unsupervised domain adaptation (DA) is to mitigate the domain shift between the source and target domains. Prior DA works show that pretext tasks could be used to mitigate this domain shift by learning domain invariant representations. However, in practice, we find that most existing pretext tasks are ineffective against other established techniques. Thus, we theoretically analyze how and when a subsidiary pretext task could be leveraged to assist the goal task of a given DA problem and develop objective subsidiary task suitability criteria. Based on this criteria, we devise a novel process of sticker intervention and cast sticker classification as a supervised subsidiary DA problem concurrent to the goal task unsupervised DA. Our approach not only improves goal task adaptation performance, but also facilitates privacy-oriented source-free DA i.e. without concurrent source-target access. Experiments on the standard Office-31, Office-Home, DomainNet, and VisDA benchmarks demonstrate our superiority for both single-source and multi-source source-free DA. Our approach also complements existing non-source-free works, achieving leading performance.
\blfootnote{* Equal contribution $|$ Webpage: \url{https://sites.google.com/view/sticker-sfda}}
\end{abstract}

\section{Introduction}
\label{sec:intro}

The prevalent trend in supervised deep learning systems is to assume that training and testing data follow the same distribution. However, such models often fail \cite{chen2017no} when deployed in a new environment (target domain) due to the discrepancy in the training (source domain) and target distributions. A standard approach to deal with this problem of \textit{domain shift} is Unsupervised Domain Adaptation (DA) \cite{ganin2016domain,CDAN}, which aims to minimize the domain discrepancy \cite{ben2006analysis} between source and target. The prime challenge in DA is to facilitate the effective utilization of the unlabeled samples while adapting to the target domain.

\begin{figure*}[t]
    \centering
    \vspace{-2mm}
    \includegraphics[width=\columnwidth]{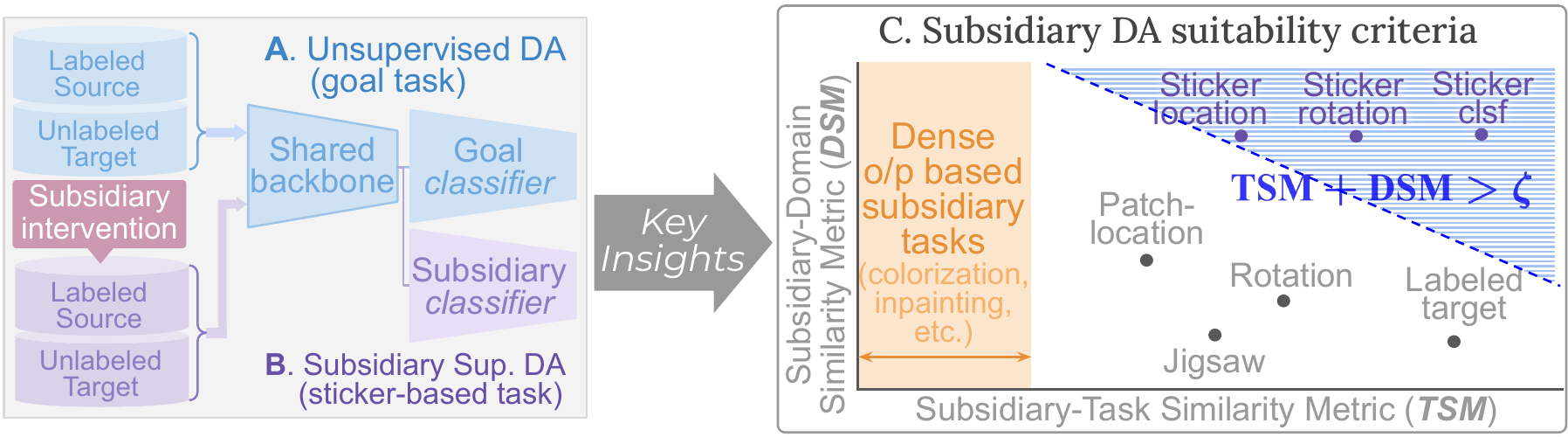}
    \vspace{-7mm}
    \caption{
    \small
    We tackle \textbf{A.} unsupervised goal task DA by introducing \textbf{B.} a concurrent subsidiary supervised DA. \textbf{C.} Our theoretical insights reveal that subsidiary tasks having both higher TSM (X-axis) and DSM (Y-axis) are most suitable for concurrent goal-subsidiary adaptation (\ie the shaded blue area). The proposed sticker-based tasks better suit concurrent goal-subsidiary DA among other self-supervised pretext tasks.
    }
    \vspace{-3mm}
    \label{fig:teaser}
\end{figure*}

Drawing motivation from self-supervised pretext task literature \cite{noroozi2016unsupervised,gidaris2018unsupervised}, recent DA works \cite{carlucci2019domain,mishra2021surprisingly} have adopted subsidiary tasks as side-objectives to improve the adaptation performance. The intuition is that subsidiary task objectives enforce learning of domain-generic representations, leading to improved domain alignment \cite{sun2019unsupervised} and consequently, better feature clustering for unlabeled target \cite{mishra2021surprisingly}. We aim to design a similar framework but, contrary to prior works, we adopt a novel perspective of subsidiary supervised DA for the subsidiary task concurrent to unsupervised goal task DA. Specifically, the framework involves a shared backbone with a goal classifier and a subsidiary classifier (Fig.\ \ref{fig:teaser}\textcolor{red}{A}, \textcolor{red}{B}).

To better understand how subsidiary supervised DA objectives support goal task DA, we intend to theoretically analyze the proposed framework. While several subsidiary tasks are available in the literature, there has been little attention on identifying the desirable properties of a subsidiary task that would better aid the unsupervised DA. A recent self-supervised work \cite{wallace2020extending} studied the effectiveness of pretraining with existing subsidiary tasks \cite{noroozi2016unsupervised,gidaris2018unsupervised} on different downstream supervised settings such as fine-grained or medical image classification \cite{parkhi2012cats,wang2017chestxray}. We argue that our intended theoretical analysis is necessary to understand the same for DA settings as DA presents a different set of challenges compared to downstream supervised learning paradigms.

% \vspace{0.2mm}
\noindent
Thus, we attempt to answer two interconnected questions,

\noindent
\textbf{1.} \textit{How does subsidiary supervised DA help goal task unsupervised DA?}

\noindent
\textbf{2.} \textit{What kind of subsidiary tasks better suit concurrent goal-subsidiary DA?}

For the first question, we uncover theoretical insights based on generalization bounds in DA \cite{ben2006analysis,zhao2019on}. 
% The generalization 
These
bounds define distribution shift or domain discrepancy between source and target as the worst discrepancy for a given hypothesis space. 
% Thus, w
We analyze the effect of adding the subsidiary supervised DA problem on the hypothesis space of the shared backbone. 
Based on this, we find that a higher domain similarity between goal and subsidiary task samples leads to a lower domain discrepancy.
This leads to better adaptation for concurrent goal-subsidiary DA w.r.t.\ naive goal DA. 
Further, we observe that a higher goal-subsidiary task similarity aids effective learning of both tasks with the shared backbone, which is crucial for subsidiary DA to positively impact the goal DA.

% For the second question, we identify the general trend in pretext task DA works (see Table \ref{tab:characteristics}) of casting the pretext task learning as a side-objective to the goal task DA objectives.
% For instance, PAC \cite{mishra2021surprisingly} uses rotation prediction while SS-DA \cite{xu2019self} uses rotated jigsaw prediction along with the DA objectives. Similarly, JiGen \cite{carlucci2019domain} used a jigsaw puzzle solving objective for domain generalization along with goal task source training. UDA-SS \cite{sun2019unsupervised} used rotation, vertical flip and patch location prediction tasks, all in parallel with the DA objectives. While these paradigms seem beneficial to the goal task DA, they overlook the domain shifts caused by the pretext tasks. We empirically verify that such pretext tasks are not domain-preserving \cite{mitsuzumi2021generalized}. Consequently, they require additional regularization techniques to support the goal task adaptation like adversarial alignment \cite{ganin2016domain} and AdaBN \cite{adabn} for SS-DA, data augmentations for JiGen, and an augmentation consistency objective for PAC. Thus, we devise a simple subsidiary DA suitability criteria to identify \textit{DA-assistive} pretext tasks by measuring domain-preservation. We also introduce a classification task based on a novel sticker intervention that follows our criteria.
% \ie, it is a DA-assistive pretext task. 
% It exhibits a positive correlation with goal task performance (Fig.\ \ref{fig:teaser}\textcolor{red}{C}) even without any additional objectives.

\begin{wrapfigure}{r}{0.68\textwidth}
    \centering
    \vspace{-7.5mm}
    \includegraphics[width=0.68\columnwidth]{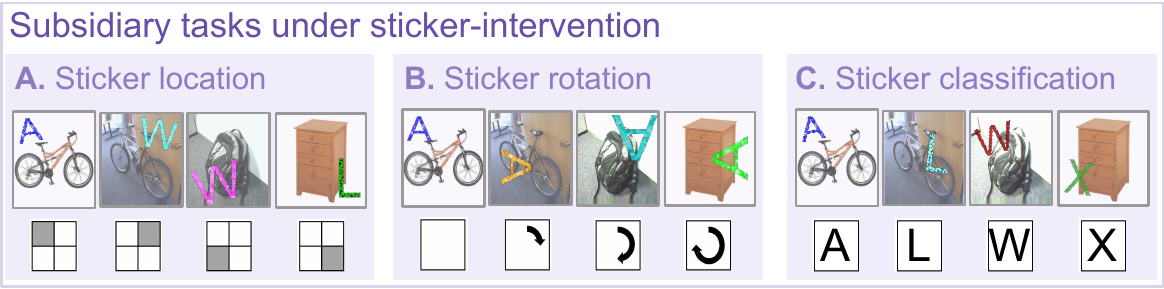}
    \vspace{-7mm}
    \caption{
    \small
    Sticker intervention involves mixup of input with a masked sticker.
    We devise the following sticker-based tasks; \textbf{A.} locating the quadrant of the sticker, \textbf{B.} predicting sticker rotation, \textbf{C.} classifying sticker category.
    }
    \vspace{-5mm}
    \label{fig:sticker_tasks}
\end{wrapfigure}

For the second question, we first devise a subsidiary-domain similarity metric (DSM) and a subsidiary-task similarity metric (TSM) to measure the domain similarity and task similarity between any subsidiary task with a given goal task. Based on our theoretical insights, we propose a subsidiary task suitability criteria using both DSM and TSM to identify \textit{DA-assistive} subsidiary tasks.
With this criteria, we evaluate the commonly used subsidiary tasks from the pretext task literature like rotation prediction \cite{mishra2021surprisingly}, patch location \cite{sun2019unsupervised}, and jigsaw permutation prediction \cite{carlucci2019domain} in Fig.\ \ref{fig:teaser}\ak{C}. We observe that these existing tasks have significantly low DSM. On the other hand, dense output based tasks like colorization \cite{larsson2017colorization} or inpainting \cite{pathak2016context} severely lack in TSM as goal task is classification-based. Understanding these limitations, we devise a sticker-intervention 
% (Fig.\ \ref{fig:sticker_intervention}) 
that facilitates domain preservation (high DSM) and propose a range of sticker-based subsidiary tasks (Fig.\ \ref{fig:sticker_tasks}). 
For general shape-based goal tasks,
it turns out that sticker classification task has the best TSM among other sticker-based tasks. This yields higher adaptation performance thereby validating the proposed criteria.

% We also observe the general trend in subsidiary task based DA works (Fig.\ \ref{fig:teaser}\ak{C}) of casting the subsidiary task learning as a side-objective to the goal DA objectives.

To evaluate our theoretical insights and the proposed concurrent subsidiary DA, we particularly focus on source-free DA regime \cite{3C-GAN,kundu2022balancing}. In this, the source and target data are not concurrently accessible while model sharing is permitted. While this challenging setting holds immense practical value by working within the data privacy regulations, we choose source-free DA as it can prominently highlight our advantages. % This is because the 
The well-developed discrepancy minimization techniques, tailored to general DA scenarios, guide the adaptation more significantly than our proposed approach but cannot be used for source-free DA.
Further, existing source-free works \cite{SHOT} rely heavily on pseudo-label based self-training on target data. Our proposed subsidiary supervised adaptation implicitly regularizes target-side self-training, leading to improved adaptation. % performance.

% \vjnote{This following paragraph is somewhat difficult to grasp in the intro. Intro with main highlights usually works better than a comprehensive intro. Instead of this paragraph, please mention some highlights in experimental results.}
% To enable source-free DA with our proposed subsidiary supervised DA, we employ out-of-distribution classification in the source domain \ie out-of-source (OOS) classification within the subsidiary task classifier. This provides domain discriminative knowledge similar to an adversarial domain discriminator \cite{ganin2015unsupervised} in non-source-free works. In the target side, we leverage this knowledge by minimizing the OOS probability to implicitly align the target domain in the absence of source data. To avoid using external data, we devise a pseudo-OOS dataset from the source data itself. 
To summarize, our main contributions are:

\vspace{-3mm}
\begin{itemize}

% \vspace{-1mm}
\item 
We introduce concurrent subsidiary supervised DA, for a subsidiary task, that not only improves unsupervised goal task DA but also facilitates source-free adaptation. We provide theoretical insights to analyze the impact of subsidiary DA on the domain discrepancy, and hence, the goal task DA.

% \vspace{-1mm}
\item 
Based on our insights, we devise a subsidiary DA suitability criteria to identify \textit{DA-assistive} subsidiary tasks that better aid the unsupervised goal task DA. We also propose novel sticker intervention based subsidiary tasks that demonstrate the efficacy of the criteria.

% \vspace{-1mm}
\item Our proposed approach achieves state-of-the-art performance on source-free single-source DA (SSDA) as well as source-free multi-source DA (MSDA) for image classification. The proposed approach also complements existing non-source-free works, achieving leading performance.

\end{itemize}

\vspace{-3mm}
\section{Related Work}
\label{sec:related_work}

% \lipsum[2]
% \lipsum[4]

% Self-supervised learning is an emerging field that applies the mechanisms of supervised learning to challenges where external supervision is not accessible. 
\vspace{-3mm}
\noindent
\textbf{Pretext tasks in self-supervised learning.} 
% The objective of SSL is to utilize unlabeled data as supervision for pretext tasks that acquire deep feature representations that are useful for downstream tasks.
Pretext tasks are used to learn deep feature representations from unlabeled data, in a self-supervised manner, for downstream tasks.
There are several pretext tasks such as image inpainting~\cite{pathak2016context}, colorization~\cite{zhang2016colorful,larsson2017colorization,zhang2017split}, spatial context prediction~\cite{doersch2015unsupervised}, contrastive predictive coding \cite{oord2018representation}, image rotation \cite{gidaris2018unsupervised}, and jigsaw puzzle solving \cite{noroozi2016unsupervised}. 
% SSL on videos \cite{wang2015unsupervised,wang2019learning} and integrating various self-supervised activities \cite{doersch2017multi} have also been investigated.
Pretext tasks are commonly used for pre-training on unlabeled data followed by finetuning on labeled data. Conversely, we perform supervised DA for the pretext-like task along with the unsupervised goal task DA, resulting in a representation that aligns the domains while maintaining the goal task performance.

\noindent
\textbf{Source-free DA.} Recently, several methods 
% for investigating source-free DA have been developed. 
have investigated source-free DA.
USFDA \cite{kundu2020universal} and FS~\cite{kundu2020towards} investigate universal DA \cite{you2019universal} and open-set DA \cite{saito2018open}, in a source-free setting by synthesizing training samples to make the decision boundaries compact.
% , allowing open classes to be recognised.
SHOT~\cite{SHOT,SHOT++}, NRC \cite{NRC} maximize mutual information and propose pseudo-labeling, using global structure to match target features to that of a fixed source classifier. To provide adaptation supervision, 3C-GAN~\cite{3C-GAN} generates labeled target-style images from a GAN. 
Finally, SFDA~\cite{liu2021source}, UR \cite{sivaprasad2021uncertainty}, and GtA \cite{kundu2021generalize} are semantic segmentation specific source-free DA techniques.
% is a segmentation method based on creating fictitious source samples.

\noindent
\textbf{Pretext task based DA.} 
Several DA works have demonstrated the efficacy of learning meaningful representations using pretext tasks. Early works \cite{ghifary2015domain,ghifary2016deep} used reconstruction as a pretext task to extract domain-invariant features. \cite{bousmalis2017unsupervised} captured both domain-specific and shared features by separating the feature space into domain-private and domain-shared spaces. \cite{carlucci2019domain} used jigsaw puzzles as a side-objective to tackle domain generalization. \cite{sun2019unsupervised} proposed that adaptation can be accomplished by learning many self-supervision tasks at the same time. \cite{kim2020cross} suggested a cross-domain SSL strategy for adaptation with minimal source labels based on instance discrimination \cite{wu2018unsupervised}. \cite{xu2019self} recommended employing SSL pretext tasks like rotation prediction and patch placement prediction. \cite{saito2020universal} solved the challenge of universal domain adaptation by unsupervised clustering. \cite{ren2018cross} employed easy labels for synthetic images, such as the surface normal, depth, and instance contour, to train a network. \cite{feng2019self} employed SSL pretext tasks like rotation prediction as part of their domain generalization technique.

\vspace{-3mm}
\section{Approach}
\label{sec:approach}

\vspace{-1mm}
In this section, we introduce required preliminaries (Sec. \ref{subsec:preliminaries}), followed by theoretical insights (Sec. \ref{subsec:theoretical}) that motivate our training algorithm design (Sec. \ref{subsec:training_algo}).

\vspace{-3mm}
\subsection{Preliminaries}
\label{subsec:preliminaries}

\vspace{-1mm}
\subsubsection{Goal task unsupervised DA.}
\begin{wrapfigure}{r}{0.3\textwidth}
    \centering
    \vspace{-8mm}
    \includegraphics[width=0.3\columnwidth]{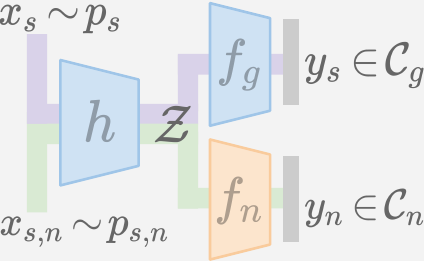}
    \vspace{-7mm}
    \caption{Our method uses a shared backbone $h$ with goal classifier $f_g$ and subsidiary classifier $f_n$.
    }
    \vspace{-6mm}
    \label{fig:fig2a}
\end{wrapfigure}

For closed set DA problem, consider a labeled source dataset $\boldsymbol{{{\mathcal{D}}_s = \{( x_s, y_s) : x_s \!\in\! {\mathcal{X}}, y_s \!\in\! {\mathcal{C}}_g\}}}$ where $\mathcal{X}$ is the input space and $\mathcal{C}_g$ denotes the label set for the goal task. $x_s$ is drawn from the marginal distribution $p_s$. Let $\boldsymbol{{\mathcal{D}}_t = \{ x_t : x_t \!\in \!{\mathcal{X}}\}}$ be an unlabeled target dataset with $x_t \!\sim\! p_t$. The goal is to assign labels for each target image $x_t$. The usual approach \cite{ganin2016domain,sun2016deepcoral,long2015learning} is to use a backbone feature extractor $h:\mathcal{X}\!\to\! \mathcal{Z}$ followed by a goal classifier $f_g: \mathcal{Z}\to \mathcal{C}_g$ (see Fig.\ \ref{fig:fig2a}). The expected source risk with $h$ and an optimal labeling function $f_S: \mathcal{X}\!\to\!\mathcal{C}_g$, is $\epsilon_s(h) = \expectation_{x\sim p_s}[\mathbbm{1}(f_g\!\circ\! h(x) \!\neq\! f_S(x))]$, where $\mathbbm(.)$ is an indicator function. Similarly, $\epsilon_t(h)$ is the target risk with optimal labeling function $f_T:\mathcal{X}\!\to\! \mathcal{C}_g$. We restate the theoretical upper bound on target risk from \cite{zhao2019on}. For backbone hypothesis $h\!\in\!\mathcal{H}$ with $\mathcal{H}$ being the hypothesis space and a domain classifier $f_d: \mathcal{Z}\!\to\!\{0, 1\}$ (0 for source, 1 for target),

{
\begingroup\makeatletter\def\f@size{9}\check@mathfonts
\def\maketag@@@#1{\hbox{\m@th\large\normalfont#1}}%
\vspace{-8mm}
\begin{align}
\epsilon_t(h) \!\leq\! &\; \epsilon_s(h) + d_\mathcal{H}(p_s, p_t)\!+\!\lambda_g; \;\;\lambda_g \!=\! 
\min\! \left\{ \expectation_{p_s}[\mathbbm{1}(f_S(x) \!\neq\! f_T(x))], \expectation_{%x \sim 
p_t}[\mathbbm{1}(f_S(x) \!\neq\! f_T(x))] \right\} \notag \\
    &\text{where, } d_\mathcal{H}(p_s,p_t)\!=\!\sup_{h \in \mathcal{H}} \left|\expectation_{x\sim p_s}[\mathbbm{1}(f_d \!\circ\! h(x) \!=\! 1)] \!-\! \expectation_{x \sim p_t}[\mathbbm{1}(f_d \!\circ\! h(x)\! =\! 1)]\right|
    \label{eqn:target_risk_bound}
\end{align} \endgroup
}
\vspace{-4mm}

%%%%%%%%%%%%% Figure Theory %%%%%%%%%%%%%%%%%%%%%%%%%%
%%%%%%%%%%%%% %%%%%%%%% %%%%%%%%%%%%%%%%%%%%%%%%%%
\noindent
% which holds for any hypothesis $h = f_g \circ h$ from a hypothesis class $\mathcal{H}$, and 
Here, $d_\mathcal{H}$ is the $\mathcal{H}$-divergence \cite{ben2006analysis} that indicates the distribution shift 
or worst-case domain discrepancy 
between the two domains.
% Specifically, $d_\mathcal{H}(p_s, p_t)=\sup_{h \in \mathcal{H}}\vert\Pr_{x \sim p_s}[f_d \circ h(x) = 1] - \Pr_{x \sim p_t}[f_d \circ h(x) = 1]\vert$, where $f_d: \mathcal{Z}\to\{0, 1\}$ is a domain classifier, $p_s$ and $p_t$ are the source and target marginal distributions, and 
% The third term involving $\min$ (denoted by $\lambda_g$ henceforth) 
$\lambda_g$ is a constant that represents the optimal cross-domain error of the labeling functions. 
% which represents the conditional distribution shift. 
Thus, the target risk $\epsilon_t(h)$ is upper bounded by these two terms along with the source risk $\epsilon_s(h)$.
% \vjnote{If possible, try to give an intuitive explanation of eq-1}
% Thus, the target risk is upper bounded by the source risk $\epsilon_s(h)$, the worst case domain discrepancy for the space $\mathcal{H}$ and the conditional distribution shift of the DA problem.

\vspace{-6mm}
% \vspace{1mm}
% \noindent
% \textbf{3.1.2. Subsidiary supervised DA.} 
\subsubsection{Subsidiary supervised DA.}
%% Subsidiary supervised DA --> How to do it? --> Pretext task. (we get labels for both source and target) define Cn before fn. Figure 2 --> show h, f_g, f_n and i/p, o/p, Z spaces. Add short eq2 similar to eq1 and use only et and not et,g where dH(s,t) is exactly same as in eq1 (mention why).
Next, we introduce a subsidiary supervised DA problem
%We propose to tackle a subsidiary supervised DA problem,
concurrent to the goal task unsupervised DA. To this end, we aim to devise a subsidiary classification task 
% by introducing 
with a new label set $\mathcal{C}_n$. The label-set specific attributes are inflicted on $x\!\in\!\mathcal{X}$ via an intervention, to form supervised pairs. These pairs form labeled source, $\boldsymbol{( x_{s,n}, y_n)\!\in\!{\mathcal{D}}_{s,n}}$ and labeled target, $\boldsymbol{( x_{t, n}, y_n)\!\in\!{\mathcal{D}}_{t, n}}$ datasets. Here, the inputs $x_{s,n}$ and $x_{t,n}$ are drawn from marginal distributions $p_{s,n}$ and $p_{t,n}$ respectively. We also define the optimal labeling functions for source and target subsidiary task as $f_{S, n}\!:\!\mathcal{X}\!\to\! \mathcal{C}_n$ and $f_{T, n}\!:\!\mathcal{X}\!\to\! \mathcal{C}_n$. 
%This yields a 
%To enable this subsidiary supervised DA, we employ a pretext task as it can produce pretext supervision for both source and target. Consider a pretext task with label set $\mathcal{C}_n$. 
Next, the prediction mapping involves the shared goal-task backbone $h$ followed by a subsidiary classifier $f_n\!:\! \mathcal{Z}\!\to\!\mathcal{C}_n$ (see Fig.\ \ref{fig:fig2a}). Here, the source-subsidiary task error is $\epsilon_{s,n}(h) = \expectation_{x\sim p_{s,n}}[\mathbbm{1}(f_n\circ h(x) \neq f_{S,n}(x))]$. Similarly, $\epsilon_{t,n}(h)$ for target and $\lambda_n$ defined as in Eq.\ \ref{eqn:target_risk_bound}. 
% Considering $p_{s,n}\approx p_s$ and $p_{t,n}\approx p_t$, we state the generalization bounds for the subsidiary DA problem as,
Thus, generalization bounds for subsidiary DA with the same $\mathcal{H}$ %as in Eq.\ \ref{eqn:target_risk_bound} 
is stated as, 
% \vjnote{$\lambda_n$ is not defined?}

\vspace{-4mm}
\begin{equation}
    \;\;\;\;\;\; \epsilon_{t, n}(h) \leq \epsilon_{s, n}(h) + d_\mathcal{H}(p_{s,n},p_{t,n}) + \lambda_n
    \label{eqn:pretext_bounds}
\end{equation}

% \noindent
% Here, $d_\mathcal{H}(p_s,p_t)$ is the same as in Eq.\ \ref{eqn:target_risk_bound} as both share the same backbone $h$.
% Backbone hypothesis space $\mathcal{H}$ is same as in Eq.\ \ref{eqn:target_risk_bound} as the backbone $h$ is shared.
% $\lambda_n$ is similar to the $\min$ term in Eq.\ \ref{eqn:target_risk_bound} where $f_S, f_T$ are replaced by $f_{S,n}, f_{T, n}$ respectively. We assume $f_g$ and $f_n$ to be frozen domain-invariant classifiers and focus only on the backbone hypothesis $h$ to be optimized for unsupervised DA.

% \ak{Introduce both the metrics for use in metrics} 

% \noindent
% \textbf{3.1.3. Domain Similarity Metric (DSM).} 

\vspace{-6mm}
% \vspace{1mm}
% \noindent
% \textbf{3.1.3. Metrics.} 
\subsubsection{Metrics.}
\label{subsubsec:metrics}
We introduce two metrics that form the basis of our insights. 

\noindent
\textbf{a) Subsidiary-Domain Similarity Metric (DSM), $\boldsymbol{\gamma_\textit{DSM}(.,.)}$.} DSM %(\eg $\gamma_\textit{DSM}(\mathcal{D}_s, \mathcal{D}_{s,n})$)
measures the similarity between two domains 
% based on 
as the inverse of
the standard $\mathcal{A}$-distance \cite{ben2006analysis}. $\mathcal{A}$-distance can be thought of as a proxy \cite{ganin2016domain} for $\mathcal{H}$-divergence.

\vspace{0.5mm}
\noindent
\textbf{b) Subsidiary-Task Similarity Metric (TSM), $\boldsymbol{\gamma_\textit{TSM}(.,.)}$.} TSM measures the task similarity of a subsidiary task w.r.t. the goal task. TSM is computed using the standard linear evaluation protocol \cite{salman2020do} borrowed from transfer learning and self-supervised literature. It is the performance of a subsidiary-task linear classifier attached to a %source-domain 
goal-task pretrained backbone feature extractor $h_{s,g}$. Intuitively, it indicates the extent of compatibility between the two tasks.
% goal and subsidiary tasks.

% We also introduce a Subsidiary-Task Suitability Metric (TSM) to measure the suitability of a subsidiary task w.r.t. the goal task. TSM is computed 
% % as the accuracy of a linear classifier $f_n$ trained for the subsidiary task with features from a frozen backbone $h_{s,g}$ (which was trained on original source data for the goal task).
% using the standard linear evaluation protocol \cite{salman2020do} from transfer learning and self-supervised learning literature. The linear evaluation is done using a source goal task pretrained backbone $h_{s,g}$ for the subsidiary task.
% Intuitively, it indicates the extent of compatibility between the goal task and subsidiary task.
% For instance, 

For a dataset pair of source-goal and source-subsidiary, \ie $(\mathcal{D}_s, \mathcal{D}_{s,n})$;

% \noindent
% DSM $\gamma_d$ and TSM $\gamma_n$ for goal source data $\mathcal{D}_s$ and subsidiary source data $\mathcal{D}_{s,n}$,

\vspace{-5mm}
\begingroup\makeatletter\def\f@size{9}\check@mathfonts
\def\maketag@@@#1{\hbox{\m@th\large\normalfont#1}}%
\begin{equation}
    % \text{\textbf{DSM}}: 
    \gamma_\textit{DSM}(\mathcal{D}_s, \mathcal{D}_{s,n}) = 1\!-\!\frac{1}{2}d_\mathcal{A}(\mathcal{D}_s, \mathcal{D}_{s,n});\;\;\; 
    % \text{\textbf{TSM}}: 
    \gamma_\textit{TSM}(\mathcal{D}_s, \mathcal{D}_{s,n}) = 1\!-\! \min_{f_n} \hat{\epsilon}_{s,n}(h_{s,g})
    \label{eqn:metrics}
\end{equation} \endgroup
\vspace{-4mm}

\noindent
Here, $d_\mathcal{A}(.,.)$ denotes $\mathcal{A}$-distance and $\hat{\epsilon}_{s,n}(.)$ denotes empirical error for subsidiary task on source data.
Note that $0 \!\leq\! \hat{\epsilon}_{s,n}(h_{s,g}) \!\leq \!1$ while $0 \!\leq\! d_\mathcal{A}(\mathcal{D}_1, \mathcal{D}_2) \!\leq\! 2$.
% \vjnote{Mention how these metrics relate to DA performance. Like whether higher the DSM/TSM the better, something like that.}

% \noindent
% \textbf{3.1.4. Subsidiary task Suitability Metric (TSM).}

\vspace{-1mm}
\subsection{Theoretical insights}
\label{subsec:theoretical}

% \noindent
% \ak{How to get $f_g$ and $f_n$? TODO - mention that we assume them to be domain-invariant classifiers.}
%% Equations

%% 3 separate subsections with one configuration each with sub-headings. Write with separate equations, not inline.

\vspace{-1mm}
% \noindent
We analyze the impact of solving subsidiary supervised DA on the goal task unsupervised DA. 
We first consider the combined bounds (combining Eq.\ \ref{eqn:target_risk_bound}, \ref{eqn:pretext_bounds}),

\vspace{-3mm}
\begin{equation}
    \epsilon_t(h)+\epsilon_{t, n}(h) \leq \epsilon_s(h)+\epsilon_{s, n}(h) + d_\mathcal{H}(p_s,p_t) 
    + d_\mathcal{H}(p_{s,n},p_{t,n}) 
    + \lambda_g+\lambda_n
    \label{eqn:combined_bounds}
\end{equation}

\noindent
% Here, the $\lambda$ terms are unaffected by the hypothesis space (or training). Since the subsidiary supervised DA affects the optimization of the shared backbone $h$, we focus on the source error terms $\epsilon_s(h)+\epsilon_{s,n}(h)$ and the domain discrepancy term $d_\mathcal{H}(p_s,p_t)$ for our analysis.
Among the six terms on the right side, the two $\lambda$ terms are constants as they do not involve the hypothesis $h$ or hypothesis space $\mathcal{H}$. We analyze 
% ways to minimize 
the source error duet, $\epsilon_s(h)+ \epsilon_{s, n}(h)$, and the domain discrepancy duet $d_\mathcal{H}(p_s,p_t)+d_\mathcal{H}(p_{s,n},p_{t,n})$.

\vspace{-5mm}
\subsubsection{Analyzing the domain discrepancy duet.\ }

First, we analyze w.r.t.\ the domain discrepancy duet by considering the following three configurations. %(shown in 
% (Fig. \ref{fig:fig2}):
% Now, we consider the following three configurations.

%%%%%%%%%%%%% Figure Theory %%%%%%%%%%%%%%%%%%%%%%%%%%
\begin{wrapfigure}{r}{0.3\textwidth}
    \centering
    \vspace{-7mm}
    \includegraphics[width=0.3\columnwidth]{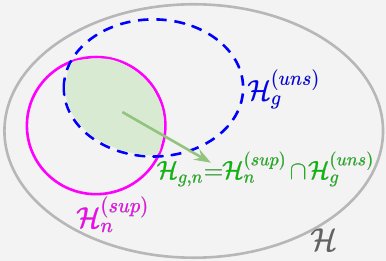}
    \vspace{-8mm}
    \caption{
    Three configurations of hypothesis spaces.
    }
    \vspace{-7mm}
    \label{fig:fig2}
\end{wrapfigure}
%%%%%%%%%%%%% %%%%%%%%% %%%%%%%%%%%%%%%%%%%%%%%%%%

\vspace{0.5mm}
\noindent
\textbf{a)} 
% While performing unsupervised adaptation only on the goal task,
While performing only unsupervised goal task DA,
the backbone optimization would operate on a limited hypothesis space $\mathcal{H}_g^{(uns)} \!\!\subset\! \mathcal{H}$ where %(see Fig. \ref{fig:fig2}\textcolor{red}{B}) boldsymbol
${\mathcal{H}_g^{(uns)} = \{ h\! \in\! \mathcal{H} : \vert \epsilon_{t}(h) - \epsilon_{s}(h)\vert \leq \zeta_{g}^{(uns)} \}}$.  
%
% \vspace{-4mm}
% \begin{equation}
%     \mathcal{H}_g^{(uns)} = \{ h \in \mathcal{H} : \vert \epsilon_{t}(h) - \epsilon_{s}(h)\vert \leq \zeta_{g}^{(uns)} \}
% \end{equation}
% % \vspace{-1mm}
%
%\noindent
Here, $\zeta_g^{(uns)}$ is a threshold on the source-target error gap.

\vspace{1mm}
\noindent \textbf{b)} While performing supervised adaptation only for subsidiary domain adaptation, the optimization would operate on a limited hypothesis space $\mathcal{H}_n^{(sup)} \!\!\subset\! \mathcal{H}$ %(Fig. \ref{fig:fig2}\textcolor{red}{B})  boldsymbol
\ie, ${\mathcal{H}_n^{(sup)} = \{ h \in \mathcal{H}  : \vert \epsilon_{t,n}(h) - \epsilon_{s,n}(h)\vert \leq \zeta_{n}^{(sup)} \}}$.
%
% \vspace{-3mm}
% \begin{equation}
%     \mathcal{H}_n^{(sup)} = \{ h \in \mathcal{H}  : \vert \epsilon_{t,n}(h) - \epsilon_{s,n}(h)\vert \leq \zeta_{n}^{(sup)} \}
% \end{equation}
%
% \noindent
\\ Here, $\zeta_{n}^{(sup)}$ is a threshold on the subsidiary-task source-target error gap.

\vspace{1mm}
\noindent \textbf{c)} While concurrently performing \textbf{a)} unsupervised goal task DA and \textbf{b)} subsidiary supervised DA (\ie the proposed approach), the optimization would operate on a limited hypothesis space $\mathcal{H}_{g,n} \!\!\subset\! \mathcal{H}$. % (see Fig. \ref{fig:fig2}\textcolor{red}{B}) boldsymbol
Specifically, ${\mathcal{H}_{g,n} \!=\! \mathcal{H}_{n}^{(sup)} \cap \mathcal{H}_g^{(uns)}}$.
%
% \vspace{-2mm}
% \begin{equation}
%     \mathcal{H}_{g,n} = \mathcal{H}_{n}^{(sup)} \cap \mathcal{H}_g^{(uns)}
% \end{equation}
%
% \noindent
This is because the backbone is shared between the two DA tasks and hence, would be limited to the intersection space.

\vspace{1mm}
Different configurations lead to different $\mathcal{H}$-spaces and consequently, different $\mathcal{H}$-divergences. Comparing the $\mathcal{H}$-divergences leads us to the following insight. %comparing the $\mathcal{H}$-divergences.

\vspace{1.5mm}
\noindent
% \textbf{Insight 1.}
\textbf{Insight 1. ($\boldsymbol{\mathcal{H}}$-divergence in concurrent goal DA and subsidiary DA)}
\textit{The backbone hypothesis space for concurrent unsupervised goal DA and subsidiary supervised DA,} \ie \textit{$\mathcal{H}_{g,n} \!=\! \mathcal{H}_{n}^{(sup)} \cap \mathcal{H}_g^{(uns)}$ will yield a lower $\mathcal{H}$-divergence than $\mathcal{H}_{g}^{(uns)}$ (hypothesis space for only unsupervised goal task DA)}, \ie

\vspace{-5mm}
\begin{equation}
    d_{\mathcal{H}_{g,n}}(p_s, p_t) \leq d_{\mathcal{H}_g^{(uns)}}(p_s, p_t) \;\;\text{and} \;\; d_{\mathcal{H}_{g,n}}(p_{s,n}, p_{t,n}) \leq d_{\mathcal{H}_g^{(uns)}}(p_{s,n}, p_{t,n})
    \label{eqn:insight1}
\end{equation}

\vspace{1mm}
\noindent \textbf{Remarks.}
% The supports for this insight are twofold.
% First, as the subsidiary supervised DA uses both source and target samples, more samples will be available for training the backbone to be domain-invariant \ie lower $d_{\mathcal{H}}$ compared to only goal task DA. Second, 
In Eq.\ \ref{eqn:target_risk_bound}, %indicates that
$d_{\mathcal{H}}(p_s, p_t)$ is the supremum over the hypothesis space $\mathcal{H}$ \ie a worst-case measure. Since $\mathcal{H}_{g,n}\!\subset\!\mathcal{H}_g^{(uns)}$, %(Fig.\ \ref{fig:fig2}\textcolor{red}{B})
$\mathcal{H}_{g,n}$
% Thus, a smaller hypothesis space $\mathcal{H}_{g,n}$ (green area in Fig. \ref{fig:fig2}) compared to $\mathcal{H}_g^{(uns)}$ 
would have a lower $\mathcal{H}$-divergence as the worst-case hypothesis of $\mathcal{H}_g^{(uns)}$ may be absent in the subset $\mathcal{H}_{g,n}$. This applies to both pairs,
% of marginal distributions, 
$(p_s, p_t)$ and $(p_{s,n}, p_{t,n})$.
While a lower $\mathcal{H}$-divergence duet leads to improved goal DA, the equality may hold %in the worst case 
when the worst hypothesis of $\mathcal{H}_g^{(uns)}$ remains in $\mathcal{H}_{g, n}$. In such a case, 
% subsidiary supervised DA would not assist the goal task DA.
concurrent DA would perform the same as naive goal DA. To this end, we put forward the following insight.

%We provide an empirical criterion to identify whether a subsidiary task could cause this issue.

\vspace{1mm}
\noindent
\textbf{Insight 2.\ 
% (Goal \& Subsidiary Domain Similarity for Concurrent DA)
{\fontsize{9.5}{9.7}\selectfont
(When is concurrent DA strictly better than naive goal DA?)}}
\textit{
% Assuming $p_s \approx p_{s,n}$ and $p_t \approx p_{t,n}$, 
A subsidiary task supports the strict inequality $d_{\mathcal{H}_{g,n}}(p_s, p_t) \!<\! d_{\mathcal{H}_g^{(uns)}}(p_s, p_t)$ if with at least $(1-\delta)$ probability, the subsidiary-domain similarity $\gamma_\textit{DSM}(\mathcal{D}_s, \mathcal{D}_{s,n})$ exceeds a threshold $\zeta_d$ by no less than $\xi$}; $\boldsymbol{\mathbb{P}[\gamma_\textit{DSM}(\mathcal{D}_s, \mathcal{D}_{s,n}) \geq \zeta_d - \xi] \geq 1-\delta}$.

% \vspace{-3mm}
% \begin{equation}
%     \mathbb{P}[\gamma_\textit{DSM}(\mathcal{D}_s, \mathcal{D}_{s,n}) \geq \zeta_d - \xi] \geq 1-\delta
%     \label{eqn:insight2}
% \end{equation}

\vspace{1mm}
\noindent
\textbf{Remarks.}
In other words, the strict inequalities in Eq.\ \ref{eqn:insight1} would hold if the DSM $\gamma_\textit{DSM}(.,.)$ exceeds a threshold $\zeta_d$.
% with the assumption that $p_s \approx p_{s,n}$ and $p_t \approx p_{t,n}$. 
The supports for this insight are twofold. First, a subsidiary task may heavily alter domain information \cite{mitsuzumi2021generalized}, \eg jigsaw shuffling \cite{carlucci2019domain}. 
Then, the backbone will be updated using out-of-domain samples which is undesirable as such samples are unlikely for inference. This will be avoided if Insight \ak{2} is satisfied. 
% Second, if $p_s \approx p_{s,n}$ and $p_t \approx p_{t,n}$, 
Second, if DSM is high, we can approximate $p_s \approx p_{s,n}$ and $p_t \approx p_{t,n}$. Thus,
more samples from subsidiary task data will be available for training the backbone to be domain-invariant (as subsidiary task uses samples from both the domains) %both source and target samples) 
\ie reducing $d_{\mathcal{H}}$ against the same in naive goal DA.

\vspace{-6mm}
\subsubsection{Analyzing the source error duet.}

Now we analyze w.r.t.\ the source error duet of Eq.\ \ref{eqn:combined_bounds}. While the $\mathcal{H}$-divergence is lower for concurrent goal task DA and subsidiary supervised DA, a logical concern is that simultaneous minimization of errors, \ie $\epsilon_s(h)+\epsilon_{s,n}(h)$, for both tasks may be difficult with the shared backbone $h$. 
% This is a more severe problem in source-free DA. The unavailability of labeled source during adaptation makes it difficult to maintain a low goal task error $\epsilon_s(h)$. Thus, we propose the following modifications. 
% to a naive subsidiary task setup.
% for source-free DA.
% \noindent
% \textbf{Modifications for source-free DA.}
% A naive source-free setup would involve simultaneous source training of the backbone $h$ and both classifiers $f_g, f_n$. To facilitate future source-free adaptation, we perform the source training in two steps (Fig.\ \ref{fig:fig3}\ak{A}). First, the backbone $h$ and goal classifier $f_g$ are trained on goal task source data $\mathcal{D}_s$. Then, the backbone $h$ is frozen and only the subsidiary classifier $f_n$ is trained on subsidiary task source data $\mathcal{D}_{s,n}$. 
% % This freezing of the backbone 
% The frozen backbone
% ensures that the goal task error $\epsilon_s(h)$ for concurrent goal-subsidiary DA (Eq.\ \ref{eqn:combined_bounds}) remains optimum \ie same as 
% % the usual case 
% without subsidiary DA (Eq.\ \ref{eqn:target_risk_bound}).
% This raises another concern where the subsidiary task source error $\epsilon_{s,n}(h)$ may be high due to the frozen backbone although goal task error may be optimum. 
Further, it may happen that simultaneous training for both tasks in target domain may hamper the goal task performance as it is unsupervised.
In such cases, the subsidiary task would be ill-equipped to assist the goal task adaptation. To avoid these, we propose another empirical criterion as follows.

\vspace{2mm}
\noindent
\textbf{Insight 3. (Goal and subsidiary task similarity for concurrent DA)}
\textit{Higher goal-subsidiary task similarity (TSM) 
% between goal and subsidiary tasks 
aids effective minimization of both task errors with the shared backbone, which is crucial for subsidiary supervised DA to positively affect the goal task DA. The criterion is $\boldsymbol{\gamma_\textit{TSM}(\mathcal{D}_s, \mathcal{D}_{s,n}) > \zeta_n}$}.

% \vspace{-3mm}
% \begin{equation}
%     \gamma_\textit{TSM}(\mathcal{D}_s, \mathcal{D}_{s,n}) > \zeta_n
%     \label{eqn:tsm_criterion}
% \end{equation}

\vspace{1mm}
\noindent
\textbf{Remarks.} Here, $\zeta_n$ is a threshold. The TSM $\gamma_\textit{TSM}$ indicates the compatibility of goal task features to support the subsidiary task. Intuitively, a higher TSM implies more overlap in the discriminative features of the two tasks, which would allow better simultaneous minimization of both task errors.

Based on Insight \ak{1}, concurrent subsidiary supervised DA and goal task DA yields a lower domain discrepancy. Further, based on Insight \ak{2}, a subsidiary task can be selected such that effective minimization of both source errors is possible simultaneously. Thus, using Eq.\ \ref{eqn:target_risk_bound}, we can infer that $\sup_{h \in \mathcal{H}_{g, n}} \epsilon_t(h) \leq \sup_{h \in \mathcal{H}_g^{(uns)}} \epsilon_t(h)$ \ie a lower target error upper bound for our approach \wrt naive goal task DA.
Now, we summarize the criteria (Insight \ak{2}, \ak{3}).

% \ak{Sec 3.2 from CVPR--> till the result (till eq 6, insight 1).
% Add DSM in insight statement.
% }

% \noindent
% \textbf{Requirements for source-free DA.}
% % \noindent
% % \textbf{Insight-2 (enabling SF).} 
% While adhering to the SF, about the sum. Freezing h, well suits the goal-error remains optimum (eq 3), against the natural case w/o subsidiary DA (eq 1).

% simultaneous has issues.. issue, 
% [para] The above raise a concern that E(s,n) could be very high?, solution--> and 

% insight-2: task-shift needs to be low (TSM metric).

% [main focus] Definition-1 (Subsidiary DA suitability criteria)

% Remarks: small (figure..)

\vspace{1.5mm}
\noindent
\textbf{Definition 1. (Subsidiary DA suitability criteria)}
% \textit{A pretext task is DA-assistive} \ie \textit{suitable for subsidiary supervised DA if the $\mathcal{H}$-divergence between the pretext task samples and original samples is less than a threshold $\zeta$,} 
\textit{A subsidiary task is termed DA-assistive} \ie \textit{suitable for subsidiary supervised DA if the sum of DSM $\gamma_\textit{DSM}$ and TSM $\gamma_\textit{TSM}$ exceeds a threshold $\zeta$,}

\vspace{-3mm}
\begin{equation}
    % d_\mathcal{H}(\mathcal{D}_s, \mathcal{D}_{s, n}) < \zeta
    % d_\mathcal{H}(p_s, p_{s, n}) < \zeta
    \gamma_\textit{DSM}(\mathcal{D}_s, \mathcal{D}_{s,n})+\gamma_\textit{TSM}(\mathcal{D}_s, \mathcal{D}_{s,n}) > \zeta
    \label{eqn:pretext_task_selection}
\end{equation}

\noindent
\textbf{Remarks.} 
% \ak{TODO - Fig.\ 1 line showing the criteria....}
In other words, a subsidiary task which is domain-preserving and has high task similarity w.r.t.\ the goal task is \textit{DA-assistive} \ie suitable for subsidiary supervised DA to aid the goal task DA. We employ this criteria empirically for a diverse set of subsidiary tasks (shown in Fig.\ \ref{fig:teaser}\ak{C}). Next, we describe the motivation for our proposed sticker intervention and corresponding subsidiary tasks as well as training algorithms tailored for source-free DA.

\vspace{-3mm}
\subsection{Sticker intervention based subsidiary task design}
\label{subsec:subsidiary_task_design}
\vspace{-2mm}

While one may consider pretext tasks from the self-supervised learning literature as candidates for subsidiary DA, almost all such tasks fail to satisfy subsidiary DA suitability criteria in Eq.\ \ref{eqn:pretext_task_selection}. For instance, dense output based tasks such as colorization \cite{zhang2016colorful,larsson2017colorization}, inpainting \cite{pathak2016context}, \etc exhibit markedly low task similarity (TSM) against the non-dense goal tasks. Further, the input intervention for certain pretext tasks such as jigsaw \cite{carlucci2019domain}, patch-location\cite{sun2019unsupervised}, rotation \cite{mishra2021surprisingly,xu2019self}, significantly alter the domain information leading to low domain similarity (DSM). 

\vspace{1mm}
\noindent
\textbf{Insight 4. (Sticker-intervention based tasks well suit subsidiary DA)} \textit{Sticker intervention is the process of pasting a sticker $x_n$ (i.e., a symbol with random texture and scale) on a given image sample $x_s\!\in\!\mathcal{D}_s$ to obtain a stickered sample, i.e.\ $x_{s,n} \!=\! \mathcal{T}(x_s, x_n) \!\in\!\mathcal{D}_{s,n}$. Following this, the subsidiary task could be defined as the classification of some sticker attribute (e.g. shape, location, or orientation). Such a formalization provides effective control to maximize $\gamma_\textit{DSM}(\mathcal{D}_s,\mathcal{D}_{s,n})$ and $\gamma_\textit{TSM}(\mathcal{D}_s,\mathcal{D}_{s,n})$, in line with our suitability criteria.}

\begin{wrapfigure}{r}{0.33\textwidth}
    \centering
    \vspace{-8.5mm}
    \includegraphics[width=0.32\columnwidth]{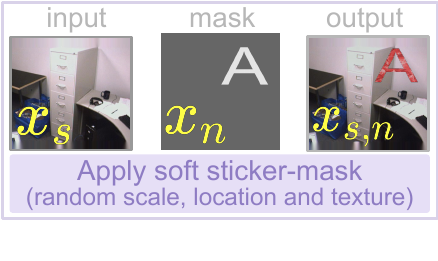}
    \vspace{-8mm}
    \caption{
    \small
    % We tackle \textbf{A.} unsupervised goal task DA by introducing \textbf{B.} a concurrent subsidiary supervised DA facilitated by sticker intervention. 
    % Performing only the subsidiary supervised DA improves the unsupervised DA performance (\textbf{C}).
    % \textbf{C.} Adapting the shared backbone by only supervised sticker DA improves goal task DA, even for \textit{source-free} DA.
    % We observe a positive correlation between the sticker and goal task performance (\textbf{C.}) even without adapting the goal task.
    %The process of 
    Sticker intervention. % involves mixup of input with a masked sticker. The mask is formed by placing a sticker (symbol with random texture) at a random location with random scale. 
    % We devise the following sticker-based tasks; \textbf{A.} locating the quadrant of the sticker, \textbf{B.} predicting sticker rotation, \textbf{C.} classifying sticker category.
    }
    \vspace{-7mm}
    \label{fig:sticker_intervention}
\end{wrapfigure}

\vspace{1mm}
\noindent
\textbf{Remarks.} The sticker intervention (Fig.\ \ref{fig:sticker_intervention}) facilitates domain preservation while simultaneously supporting a range of subsidiary tasks. Since the proposed sticker intervention alters only a local area of the sample, the original content is not suppressed which in turn preserves the domain information, implying high DSM. Following this, one can ablate over a range of sticker-based tasks in order to select a suitable subsidiary task based on the given goal task. Below, we discuss some possible subsidiary tasks under the sticker intervention.

\noindent
\textbf{a) Sticker location} (Fig.\ \ref{fig:sticker_tasks}\ak{A}). We draw motivation from patch-location \cite{sun2019unsupervised}, where the task is to classify the quadrant to which a patch-input belongs. 
With sticker intervened images, the task is to classify the quadrant with the sticker. Our use of whole images as input is more domain-preserving than patch-input.

\noindent
\textbf{b) Sticker rotation} (Fig.\ \ref{fig:sticker_tasks}\ak{B}). Motivated by the image rotation task \cite{mishra2021surprisingly}, we propose sticker rotation task where the rotation of the sticker has to be classified (0\degree, 90\degree, 180\degree and 270\degree rotations possible). Note that our sticker rotation does not affect the domain information while rotating the entire image does.

\noindent
\textbf{c) Sticker classification} (Fig.\ \ref{fig:sticker_tasks}\ak{C}). While the discriminative features in the previous two tasks were location and rotation, we propose sticker classification task with primary discriminative features as shape. In other words, the task is to classify the sticker shape (\ie the symbol) given a stickered sample.

% \noindent
% \ak{---------------------------------------------- Akshay ----------------------------------------------}

\vspace{-4mm}
\subsection{Training algorithm design under source-free constraints}
\label{subsec:training_algo}

% ********************************************************
% 3.3. Implementing based in the insights 
% Execution insights 1. Sticker..
% Possible task configurations:
% a) 
% b)
% c)

% Execution insights 2. OOS

% % Inline with XX, [skip eq 11.] [don't talk about the extra implementation details, remove entirely.] --> simple ending, after eqn 11, we also use the general loss terms inline with XX.

% ********************************************************

For the standard DA setting with concurrent access to source and target data \cite{ganin2016domain,sun2016deepcoral}, the subsidiary supervised DA can be implemented simply by optimizing the subsidiary classification loss simultaneously for source and target. This would yield a lower domain discrepancy as discussed in Sec.\ \ref{subsec:theoretical}. However, in the more practical source-free setting \cite{3C-GAN,kundu2020universal} where concurrent source-target access is prohibited, this simple approach would not be possible.
% \vspace{1mm}
% \noindent
% \textbf{3.4.1. Why source-free DA?}
% % \ak{TODO - 2nd last para of intro has the arguments.} 
% % Working within the data privacy regulations, the source free DA setting contains enormous practical value, which can illustrate the benefits of our subsidiary task. 
% The source-free DA regime is immensely practical as it respects data privacy regulations.
% Besides this, the source-free setting presents new challenges which highlight the advantages of our proposed method more prominently.
% This is because, the performance in non-source-free DA is strongly influenced by well-developed discrepancy minimization techniques.
% On the other hand, these techniques cannot be leveraged in a source-free setting due to their requirement of concurrent source-target data access. Thus, we primarily operate in the source-free regime to evaluate our theoretical insights and the proposed introduction of a concurrent subsidiary supervised DA problem.
% advent of source-free techniques...
We believe the improvements will be prominent in source-free DA based on the following insight:

\vspace{1mm}
\noindent
%{\fontsize{9.5}{9.7}\selectfont
\textbf{Insight 5. (Subsidiary DA better suits challenging source-free DA).} \textit{
Existing source-free DA works heavily rely on pseudo-label or clustering based self-training on unlabeled target with no obvious alternative. The proposed subsidiary supervised adaptation helps to implicitly regularize the target-side self-training, leading to improved adaptation performance. 
The subsidiary DA not only aids goal DA as a result of high DSM but also preserves the goal task inductive bias as a result of high TSM, while adhering to the source-free constraints.
}

\vspace{1mm}
\noindent
\textbf{Remarks.} 
% The source-free DA regime is immensely practical as it respects data privacy regulations.
% Besides this, t
% The subsidiary supervised adaptation not only aids goal DA as a result of high DSM but also preserves the goal task inductive bias as a result of high TSM, while adhering to the source-free constraints.
% Further, t
% Moreover, t
The source-free setting presents new challenges which highlight the advantages of our proposed method more prominently.
This is because, the performance in non-source-free DA is strongly influenced by well-developed discrepancy minimization techniques.
However, these techniques cannot be leveraged in a source-free setting due to their requirement of concurrent source-target data access. 
Thus, we primarily operate in the source-free regime to evaluate our theoretical insights and the proposed 
% introduction of a 
concurrent subsidiary supervised DA problem.

%%%%%%%%%%%%% Figure Approach %%%%%%%%%%%%%%%%%%%%%%%%%%
% \begin{figure*}[t]
\begin{wrapfigure}{r}{0.47\textwidth}
    \centering
    \vspace{-7mm}
    \includegraphics[width=0.47\textwidth]{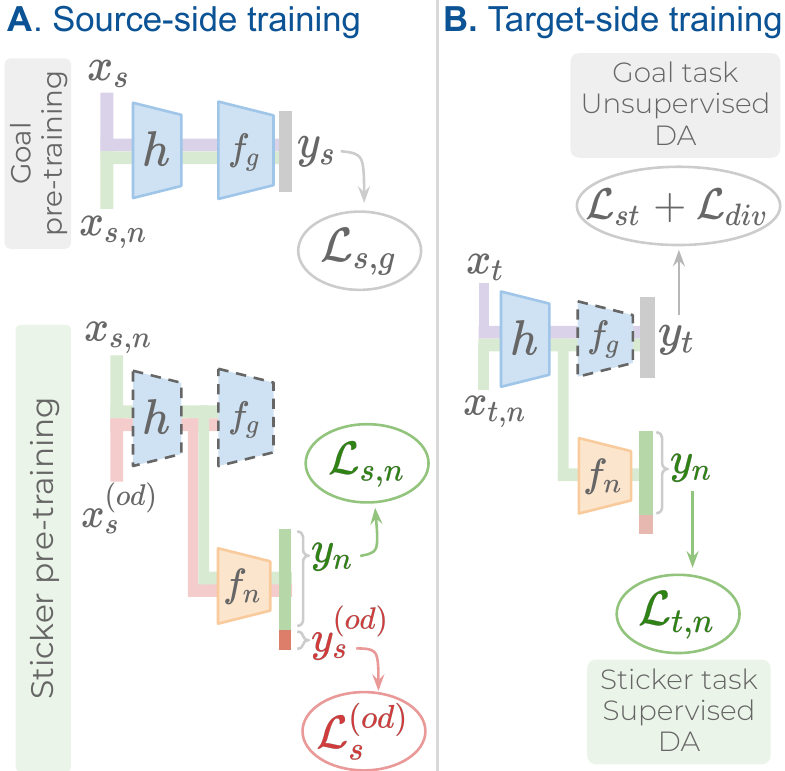}
    \vspace{-7mm}
    \caption{
    \small
    % \textbf{A.} The given labeled source $\mathcal{D}_s$ and unlabeled target $\mathcal{D}_t$ datasets. \textbf{B.} The sticker dataset $\mathcal{D}_n$ with the sticker intervention yields stickered-source $\mathcal{D}_{s, n}$ and stickered-target $\mathcal{D}_{t, n}$ datasets. The pseudo-OOS dataset $\mathcal{D}_s^{(od)}$ contains patch-shuffled source samples. Green circles only highlight the stickers and are not part of the samples. 
    \textbf{A.} Source-side training involves goal pre-training (Sec.\ \ref{subsec:training_algo}\textcolor{red}{.1}) and sticker pre-training (Sec.\ \ref{subsec:training_algo}\textcolor{red}{.2}). \textbf{B.} Target-side training involves concurrent goal-task unsupervised DA and sticker-task supervised DA (Sec.\ \ref{subsec:training_algo}\textcolor{red}{.3}).
    }
    \vspace{-7mm}
    \label{fig:fig3}    
\end{wrapfigure}
% \end{figure*}
%%%%%%%%%%%%% %%%%%%%%% %%%%%%%%%%%%%%%%%%%%%%%%%%

\vspace{1mm}
% \noindent
% \textbf{Enabling source-free DA.} 
We perform the training in three steps. First two steps involve pre-training of goal task and subsidiary task respectively with source data. The final step involves adapting both tasks to target domain. For clarity, we first summarize available and intervened datasets required for training and their notations.

%\vspace{1mm}
\noindent
\textbf{Datasets.} The goal task source data is denoted by $(x_s, y_s) \in \mathcal{D}_s$ while the corresponding unlabeled target is denoted by $x_t \in \mathcal{D}_t$. 
% (Fig.\ \ref{fig:fig3}\textcolor{red}{A}). 
%The dataset for sticker intervention, with corresponding sticker labels, is denoted by $(x_n, y_n)\in \mathcal{D}_n$.
% (Fig.\ \ref{fig:fig3}\textcolor{red}{B}).
The intervened stickered-source data, coupled with both goal and sticker task labels, is denoted by $(x_{s,n}, y_s, y_n) \in \mathcal{D}_{s,n}$. The corresponding stickered-target data, with only subsidiary sticker task labels, is denoted by $(x_{t,n}, y_n) \in \mathcal{D}_{t,n}$. 
% (Fig.\ \ref{fig:fig3}\textcolor{red}{B}). 
We introduce a pseudo-OOS (out-of-source) dataset, $\mathcal{D}_s^{\textit{(od)}}$ further in this section.

\vspace{1.5mm}
\noindent
\textbf{3.4.1. Goal task source pre-training} (Fig.\ \ref{fig:fig3}\textcolor{red}{A}). We train the backbone $h$ and goal classifier $f_g$ with source data $\mathcal{D}_s$ and stickered-source data $\mathcal{D}_{s,n}$: %The training objective is,
% For an input $x$ and goal label $y$, $\mathcal{L}_{s, g} = \mathcal{L}_{CE}(f_g\! \circ\! h(x), y)$, with the overall objective,

\vspace{-3mm}
\begin{equation}
    \min_{\theta_{h}, \theta_{f_g}} \expectation_{(x, y)\in \mathcal{D}_s \cup \mathcal{D}_{s,n}} \!\![\mathcal{L}_{s, g}] ;\;\;
    % \text{ where } 
    \mathcal{L}_{s, g} = \mathcal{L}_{ce}(f_g\! \circ\! h(x), y)
\end{equation}
% \vspace{-2mm}

\noindent
Here, $\theta_h$ and $\theta_{f_g}$ are the parameters of $h$ and $f_g$, $\mathcal{L}_{ce}$ is the cross-entropy loss, $y$ is the goal task label, and expectation is implemented by sampling mini-batches.

\vspace{1mm}
\noindent
\textbf{3.4.2. Sticker task source pre-training} (Fig.\ \ref{fig:fig3}\textcolor{red}{A}).
We pretrain the sticker classifier $f_n$ while inculcating the ability to reject samples out of the source distribution. 
% This domain discriminatory knowledge will support the future source-free target alignment. 
Specifically, $f_n$ predicts a $(\vert\mathcal{C}_n\vert +1)$-sized vector and is trained to classify \textit{out-of-source} (OOS) samples to the $(\vert\mathcal{C}_n\vert +1)^\text{th}$ class. 

\vspace{1mm}
\noindent
\textbf{Insight 6.} 
% \ak{TODO - about using OOS in source-side and target-side.}
\textit{The OOS node in the sticker classifier implicitly behaves as a domain discriminator from adversarial alignment methods. Minimizing the OOS probability only for the target data aligns the target with the source.}

\vspace{1mm}
\noindent
\textbf{Remarks.}
% \ak{TODO} 
% \lipsum[11]
In source training, the OOS objective forces the sticker classifier to discriminate between source and OOS samples. This is done with the intuition that OOS samples simulate the role of target samples in adversarial alignment methods. This domain discriminatory knowledge will support future source-free target alignment. Concretely, the shared backbone can be adapted to the target, by minimizing OOS probability for target samples, as source knowledge is preserved 
% in the frozen 
by freezing
$f_g$.
Thus, we require OOS data to prepare $f_n$ for adaptation.

\vspace{1mm}
\noindent \textbf{Obtaining the OOS dataset.}
The naive approach is to use a dataset unrelated to the goal task label set. Conversely, we devise a pseudo-OOS dataset using only the already available source samples.
Mitsuzumi \etal \cite{mitsuzumi2021generalized} show that, beyond a certain grid size, shuffling the grid patches makes the domain unrecognizable. Inspired by this, we generate the pseudo-OOS dataset by shuffling the grid patches of source images. 
% (Fig.\ \ref{fig:fig3}\textcolor{red}{B}). 
% See Suppl. for more details of OOS sample generation. 
% Concretely, we perform the patch-shuffling on the source images to obtain the pseudo-OOS dataset. 
We also perform the sticker intervention on shuffled images, at random, to further instill the differences between source and pseudo-OOS samples (see Suppl). Formally, $(x_s^{(od)}, y_s^{(od)}) \in \mathcal{D}_s^{(od)}$ where $y_s^{(od)}$ denotes the OOS category \ie the $(\vert\mathcal{C}_n\vert +1)^\text{th}$ category of $f_n$.

We train only the sticker classifier $f_n$, keeping backbone $h$ and goal classifier $f_g$ frozen, using cross-entropy loss $\mathcal{L}_{ce}$.
% The loss for the stickered source input $x_{s, n}$ with label $y_n$ is $\mathcal{L}_{s, n} = \mathcal{L}_{CE}(f_n\!\circ\! h(x_{s, n}), y_n)$. Similarly, the loss for pseudo-OOS input $x_{s}^{(od)}$ with OOS label $y_s^{(od)}$ is $\mathcal{L}_{s}^{(od)}=\mathcal{L}_{CE}(f_n\!\circ\! h(x_{s}^{(od)}), y_s^{(od)})$.
With $\mathcal{L}_{s,n} \!=\! \mathcal{L}_{ce}(f_n\!\circ\! h(x_{s,n}), y_n)$, %and $\mathcal{L}_{s}^{(od)}\!=\!\mathcal{L}_{ce}(f_n\!\circ\! h(x_{s}^{(od)}), y_s^{(od)})$
the overall objective for stickered source data $\mathcal{D}_{s,n}$ and pseudo-OOS data $\mathcal{D}_s^{(od)}$ is,

% \vspace{-5mm}
% \begin{align}
% \begin{split}
%     &\min_{\theta_{f_n}} \expectation_{\mathcal{D}_{s, n}}[\mathcal{L}_{s, n}] \text{;  } \mathcal{L}_{s, n} \!=\! \mathcal{L}_{ce}(f_n\!\circ\! h(x_{s, n}), y_n) \\
%     &\min_{\theta_{f_n}}  \expectation_{\mathcal{D}_s^{(od)}}[\mathcal{L}_{s}^{(od)}] \text{;  } \mathcal{L}_{s}^{(od)}\!=\!\mathcal{L}_{ce}(f_n\!\circ\! h(x_{s}^{(od)}), y_s^{(od)})
% \end{split}
% \end{align}

\vspace{-4mm}
\begin{equation}
    \min_{\theta_{f_n}} \expectation_{\mathcal{D}_{s,n}}[\mathcal{L}_{s,n}]\!+\!\expectation_{\mathcal{D}_s^{(od)}}[\mathcal{L}_{s}^{(od)}]; \;\;\;\text{where}\;\; 
    \mathcal{L}_{s}^{(od)}\!=\!\mathcal{L}_{ce}(f_n\!\circ\! h(x_{s}^{(od)}), y_s^{(od)})
\end{equation}
\vspace{-2mm}

%where $\mathcal{L}_{s,n} \!=\! \mathcal{L}_{ce}(f_n\!\circ\! h(x_{s,n}), y_n)$ and $\mathcal{L}_{s}^{(od)}\!=\!\mathcal{L}_{ce}(f_n\!\circ\! h(x_{s}^{(od)}), y_s^{(od)})$.

% \vspace{-2mm}
\vspace{1mm}
\noindent
\textbf{3.4.3. Source-free target adaptation} (Fig.\ \ref{fig:fig3}\textcolor{red}{B}).
For unsupervised goal task adaptation, we 
% adopt a simplified form of the 
use the general
% neighborhood clustering loss $\mathcal{L}_{nc}$ \cite{saito2020universal}
self training loss $\mathcal{L}_{st}$ and diversity loss $\mathcal{L}_{div}$ \cite{SHOT}.
% to avoid collapsing to trivial solutions. 
% The neighborhood clustering loss brings the backbone features $h(x)$ for target samples closer to its nearest neighbor using a target feature memory bank $F_t^{(mb)}$. The diversity loss ensures that goal task output $f_g \circ h(x)$ does not collapse to trivial solutions.
% \ak{One line each about the 2 losses, what are the inputs, what is the logic?}
See Suppl. for more details.
% on $\mathcal{L}_{nc}$ and $\mathcal{L}_{div}$. 
The goal task objective is given in Eq.\ \ref{eqn:target_side_obj} (left),

\vspace{-3mm}
\begingroup\makeatletter\def\f@size{9.4}\check@mathfonts
\def\maketag@@@#1{\hbox{\m@th\large\normalfont#1}}%
\begin{equation}
    % \min_{\theta_{h}} \expectation_{(x, y) \in \mathcal{D}_{t} \cup \mathcal{D}_{t, n}} [\mathcal{L}_{nc}(h(x), F_t^{(mb)}) + \mathcal{L}_{div}(f_g\!\circ\!h(x))]
    \min_{\theta_{h}} \expectation_{%(x, y) \in
    \mathcal{D}_{t} \cup \mathcal{D}_{t,n}} [\mathcal{L}_{st} + \mathcal{L}_{div}]
    \;\;\;\;\text{and}\;\;\;\;
    \min_{(\theta_{h}, \theta_{f_n})} \expectation_{\mathcal{D}_{t,n}} [\mathcal{L}_{t,n}] ; \;\;
    % \text{ where } 
    \mathcal{L}_{t,n}\!=\!\mathcal{L}_{ce}(f_n \!\circ\! h(x_{t,n}), y_n)
    \label{eqn:target_side_obj}
\end{equation} \endgroup

% Note that o
\noindent
% Only backbone $h$ is updated to preserve the goal task knowledge of the classifier $f_g$ and the objective is optimized for both original and stickered target samples.
The goal classifier $f_g$ is frozen to preserve its inductive bias and only the backbone $h$ is updated for both original and stickered samples in Eq.\ \ref{eqn:target_side_obj} (left).

\vspace{0.5mm}
For subsidiary supervised sticker adaptation, we use a simple cross-entropy loss with sticker labels.
% are available. 
We implicitly minimize OOS probability by maximizing label class probability. We observe that this works well and explicit minimization of OOS probability is not required. 
As per Insight \ak{6}, \textit{out-of-target} (OOT) samples are not required. Further, using OOT samples to update the backbone could be undesirable as discussed 
% while formulating Definition \textcolor{red}{1}.
under Insight \ak{2}.
The objective is given in Eq.\ \ref{eqn:target_side_obj} (right).
% \vspace{-3mm}
% \begin{equation}
%     \min_{(\theta_{h}, \theta_{f_n})} \expectation_{\mathcal{D}_{t,n}} [\mathcal{L}_{t,n}] ; \;\;
%     % \text{ where } 
%     \mathcal{L}_{t,n}\!=\!\mathcal{L}_{ce}(f_n \!\circ\! h(x_{t,n}), y_n)
% \end{equation}
% % \vspace{-2mm}
% Note that b
% \noindent
Both backbone $h$ and sticker classifier $f_n$ are updated as the 
% sticker 
task is supervised.

\begin{table*}[t]
    %\vspace{1mm}
    % \small
    \centering
    \setlength{\tabcolsep}{1.4pt}
    \renewcommand{\arraystretch}{0.95}
    \caption{Single-Source Domain Adaptation (SSDA) on Office-Home benchmarks. SF indicates \textit{source-free} adaptation.}
    \label{tab:ssda_office_home}
    \vspace{-3mm}
    \resizebox{1\linewidth}{!}{%
        \begin{tabular}{lccccccccccccccc}
            \toprule
%             Method & Ar$\rightarrow$Cl & Ar$\rightarrow$Pr & Ar$\rightarrow$Rw & Cl$\rightarrow$Ar & Cl$\rightarrow$Pr & Cl$\rightarrow$Rw & Pr$\rightarrow$Ar & Pr$\rightarrow$Cl & Pr$\rightarrow$Rw & Rw$\rightarrow$Ar & Rw$\rightarrow$Cl & Rw$\rightarrow$Pr & \textbf{Avg} \\
% 			\midrule
            \multirow{2}{30pt}{\centering Method} & \multirow{2}{*}{\centering SF} & \multicolumn{13}{c}{\textbf{Office-Home}} \\
            \cmidrule{4-16}
            && & {\scriptsize {Ar}$\shortto${Cl}} & {\scriptsize {Ar}$\shortto${Pr}} & {\scriptsize{Ar}$\shortto${Rw}} & {\scriptsize{Cl}$\shortto${Ar}} & {\scriptsize{Cl}$\shortto${Pr}} & {\scriptsize{Cl}$\shortto${Rw}} & {\scriptsize{Pr}$\shortto${Ar}} & {\scriptsize{Pr}$\shortto${Cl}} & {\scriptsize{Pr}$\shortto${Rw}} & {\scriptsize{Rw}$\shortto${Ar}} & {\scriptsize{Rw}$\shortto${Cl}} & {\scriptsize{Rw}$\shortto${Pr}} & Avg \\
            \midrule
			FixBi~\cite{FixBi} &\xmark& & 58.1 &77.3 &80.4 &67.7 &79.5 &78.1 &65.8 &57.9 &81.7 &76.4 &62.9 &86.7 &72.7 \\
			SENTRY\cite{SENTRY} & \xmark && 61.8 & 77.4 & 80.1 & 66.3 & 71.6 & 74.7 & 66.8 & 63.0 & 80.9 & 74.0 & 66.3 & 84.1 & 72.2 \\ % ICCV’21
			SCDA~\cite{SCDA} & \xmark && 60.7 & 76.4 & 82.8 & 69.8 & 77.5 & 78.4 & 68.9 & 59.0 & 82.7 & 74.9 & 61.8 & 84.5 & 73.1 \\ % ICCV’21
			\midrule
			
	        SHOT~\cite{SHOT} &\cmark& &{57.1} & {78.1} & 81.5 & {68.0} & {78.2} & {78.1} & {67.4} & 54.9 & {82.2} & 73.3 & 58.8 & {84.3} & {71.8}  \\
	        ${\text{A}^{2}\text{Net}}$~\cite{A2Net} &\cmark& &  58.4 & 79.0 & {82.4} & 67.5 & 79.3 & 78.9 & {68.0} & 56.2 & 82.9 & {74.1} & 60.5 & 85.0 & 72.8 \\
		    GSFDA~\cite{GSFDA} & \cmark && 57.9 & 78.6 & 81.0 & 66.7 & 77.2 & 77.2 & 65.6 & 56.0 & 82.2 & 72.0 & 57.8 & 83.4 & 71.3 \\ % ICCV’21
		    CPGA~\cite{CPGA} & \cmark && {59.3} & 78.1 & 79.8 & 65.4 & 75.5 & 76.4 & 65.7 & \textbf{58.0} & 81.0 & 72.0 & {64.4} & 83.3 & 71.6 \\ %  IJCAI’21
	        {NRC}~\cite{NRC} &\cmark& &{57.7} 	& {80.3} 	&{82.0} 	&{68.1} 	&{79.8} 	&{78.6} 	&{65.3} 	&{56.4} 	&{83.0} 	&71.0	&{58.6} &{85.6} 	&{72.2} \\
	        SHOT++\cite{SHOT++} & \cmark && 57.9 & 79.7 & 82.5 & {68.5} & 79.6 & 79.3 & {68.5} & 57.0 & \textbf{83.0} & 73.7 & 60.7 & 84.9 & 73.0 \\ %
		    
		    \rowcolor{gray!10} \textit{Ours}    &\cmark& & \textbf{61.0}	& \textbf{80.4}	& \textbf{82.5}	& \textbf{69.1}	&\textbf{79.9}	&\textbf{79.5}	& \textbf{69.1}	& {57.8}	&{82.7}	&\textbf{74.5}	& \textbf{65.1}	&\textbf{86.4} &\textbf{74.0}\\

            \bottomrule
            \end{tabular} 
        }
\end{table*}
%%%%%%%%%%%%%%%%%%%%%%%%%%%%% SSDA Office-Home Table ends. %%%%%%%%%%%%%%%%%%%%%%%%%%

\vspace{-3mm}
\section{Experiments}
\label{sec:experiments}

\vspace{-2mm}
We provide the implementation details of our experiments and thoroughly evaluate our approach w.r.t.\ {state-of-the-art} prior works across multiple settings.
Unless mentioned, \textit{Ours} implies sticker classification as the subsidiary task.

\vspace{-3mm}
\subsection{Experimental setup}
\label{subsec:setup}

\vspace{-1mm}
\noindent
\textbf{Datasets.}
% We experiment on four Domain Adaptation benchmarks to test the effectiveness of our method. 
We evaluate the effectiveness of our approach on four standard DA benchmarks.
\textbf{Office-31}~\cite{office-31} benchmark consist of three domains under office environments: Amazon (\textbf{A}), DSLR (\textbf{D}), and Webcam (\textbf{W}), each with 31 object categories.
\textbf{Office-Home}~\cite{office-home} is a more challenging dataset. It comprises of images of commonplace objects divided into four domains: Artistic (\textbf{Ar}), Clipart (\textbf{Cl}), Product (\textbf{Pr}), and Real-World (\textbf{Rw}), each with 65 classes. 
% \textbf{VisDA}~\cite{visda} is a large-scale dataset for domain adaptation from synthetic to real.
\textbf{VisDA}~\cite{visda} is a large-scale dataset for synthetic-to-real domain adaptation.
The source domain has 152,397 synthetic images, while the target domain has 55,388 real-world images.
% Finally, 
\textbf{DomainNet}~\cite{domainnet} is the most challenging due to its highly diverse domains and huge class imbalance. It has 6 domains: Clipart (\textbf{C}), Real (\textbf{R}), Infograph (\textbf{I}), Painting (\textbf{P}), Sketch (\textbf{S}) and Quickdraw (\textbf{Q}) with 345 classes each.

\vspace{1mm}
\noindent
\textbf{Implementation details.} 
We use a ResNet-50~\cite{he2016deep_resnet} backbone for Office-Home, Office-31 and DomainNet, and ResNet-101 for VisDA, for a fair comparison with prior works. We employ the same network design as SHOT \cite{SHOT}, \ie replacing the classifier with a fully connected layer with batch norm \cite{ioffe2015batch} and another fully connected layer with weight normalization \cite{salimans2016weight}. For the subsidiary classifier, we use the same architecture after ResLayer-3. The number of sticker classes is 10.
See Suppl.\ for more details related to sticker intervention and sticker-based tasks.
Following \cite{DECISION,SHOT}, we use label smoothing for source training using Adam \cite{kingma2014adam} with learning rate 1e-3, momentum 0.9, and batch size 64.
% For Office-31 and OfficeHome, we train 30 epochs, 15 for. 
We use separate Adam optimizers for each loss term to avoid loss balancing hyperparameters.

% The learning rate for Office-31, Office-Home, and DomainNet is set to 1e-3 for all layers, with the exception of the last two newly introduced fc layers, which use 1e-2. Learning rates are set 10 times lower for VisDA. 
% For Office-31 and OfficeHome, we train 30 epochs, 15 for VisDA, and 20 for DomainNet. 
% For the number of nearest neighbors we use 3 for Office-31, and Office-Home. It is set to 5 for VisDA and DomainNet due to their larger size. 
% Experiments are carried out on an Nvidia Titan Xp GPU.

% %%%%%%%%%%%%%%%%%%%%%%%%%%%%% MSDA DomainNet Table start. %%%%%%%%%%%%%%%%%%%%%%%%%%

\begin{table*}[t]
    %\vspace{1mm}
    % \small
    %\vspace{-1mm}
    \centering
    \caption{Multi-Source Domain Adaptation (MSDA) on DomainNet and Office-Home. 
    % The middle section compares single-source DA works by adapting for each source-target pair and reporting the best and worst results for each target. 
    % The top and bottom sections contain non-source-free and source-free (SF) MSDA methods respectively. 
    % SF indicates \textit{source-free} adaptation.
    We outperform \textit{source-free} (denoted by SF) prior arts despite not using domain labels.
    }
    \vspace{-3mm}
    \label{tab:msda_compare}
    \setlength{\tabcolsep}{2.0pt}
    \renewcommand{\arraystretch}{0.95}
    \resizebox{1\linewidth}{!}{%
        \begin{tabular}{lcccccccccccccccc}
            \toprule
            
            % Method & SF& & Domain Net
            \multirow{2}{30pt}{\centering Method} & \multirow{2}{*}{\centering SF} & \multirow{2}{*}{\parbox{2cm}{\centering \vspace{4pt} w/o Domain \\ Labels}} & \multicolumn{7}{c}{\textbf{DomainNet}} && \multicolumn{5}{c}{\textbf{Office-Home}} \\
            \cmidrule(lr){4-10} \cmidrule(lr){12-16}
            & & &$\rightarrow$C & $\rightarrow$I & $\rightarrow$P & $\rightarrow$Q & $\rightarrow$R & $\rightarrow$S & Avg & & $\rightarrow$Ar & $\rightarrow$Cl & $\rightarrow$Pr & $\rightarrow$Rw & Avg \\
            \midrule
            {WAMDA \cite{WAMDA}} & \xmark & \xmark &  59.3 & 21.8 & 52.1 & 9.5 & 65.0 & 47.7 & 42.6 && 71.9 & 61.4 & 84.1 & 82.3 & 74.9 \\ %, BMVC'20
            {SImpAl$_{50}$~\cite{SImpAl}} & \xmark & \xmark &   66.4 & 26.5 & 56.6 & 18.9 & 68.0 & 55.5 & 48.6 & & 70.8 & 56.3 & 80.2 & 81.5 & 72.2 \\ % , NeurIPS2020
            {CMSDA~\cite{CMSDA}} & \xmark & \xmark &  70.9 & 26.5 & 57.5 & 21.3 & 68.1 & 59.4 & 50.4 & &  71.5 & 67.7 & 84.1 & 82.9 & 76.6 \\ % , BMVC'21
            {DRT \cite{DRT}} & \xmark & \xmark &   71.0 & 31.6 & 61.0 & 12.3 & 71.4 & 60.7 & 51.3 &  & - & - & - & - & -\\ % , CVPR'21
            STEM~\cite{STEM} & \xmark & \xmark & 72.0 & 28.2 & 61.5 & 25.7 & 72.6 & 60.2 & 53.4 & & - & - & - & - & -  \\ % ICCV'21            
            \midrule
            {Source-combine} &\xmark & \cmark &57.0 &23.4 &54.1 &14.6 &67.2 &50.3 & 44.4 & & 58.0 & 57.3 & 74.2 & 77.9 & 66.9\\
            SHOT~\cite{SHOT}-Ens & \cmark & \xmark &58.6	&25.2	&55.3	&15.3	&\textbf{70.5}	&52.4	&46.2 & & 72.2 &59.3 &82.8 &82.9 &74.3\\
            DECISION~\cite{DECISION} & \cmark & \xmark &61.5	 &21.6	 &54.6 	 &\textbf{18.9}	 &67.5	&51.0	 &45.9 & & {74.5} &{59.4} &{84.4} & {83.6} &{75.5}\\
            SHOT++~\cite{SHOT++} & \cmark & \xmark & - & - & - & - & - & - & - & & 73.1 & 61.3 & 84.3 & 84.0 & 75.7 \\
            CAiDA~\cite{caida} &\cmark & \xmark & - & - & - & - & - & - & - & & \textbf{75.2} & 60.5 & {84.7} & 84.2 & 76.2\\
            NRC~\cite{NRC} & \cmark & \cmark & 65.8 & 24.1 & 56.0 & 16.0 & 69.2 & 53.4 & 47.4 & & 70.6 & 60.0 & 84.6 & 83.5 & 74.7 \\
            \rowcolor{gray!10} \textit{Ours} & \cmark & \cmark &\textbf{70.3} &\textbf{25.7} &\textbf{57.3} &{17.1} & {69.9} & \textbf{57.1} & \textbf{49.6} & &{75.1} &\textbf{64.1} &\textbf{86.6} &\textbf{84.4} & \textbf{77.6}\\ 
            \bottomrule
            \end{tabular} 
        }
\vspace{-3mm}
\end{table*}
%%%%%%%%%%%%%%%%%%%%%%%%%%%%% MSDA DomainNet Table end. %%%%%%%%%%%%%%%%%%%%%%%%%%

\vspace{-3mm}
\subsection{Discussion}
\label{sec:discussion}

\vspace{-1mm}
We provide an extensive ablation study of both the \textit{source-side} and \textit{target-side} training. Further, we show that our approach is compatible with existing non-source-free DA works and achieves faster and improved convergence.

\vspace{-3mm}
\subsubsection{Comparison with prior arts}
\label{subsec:sota_compare}

\hfill\\

\vspace{-4mm}
\noindent
\textbf{a) Single Source Domain Adaptation (SSDA).} 
We compare our proposed approach with prior source-free SSDA works in Table \ref{tab:ssda_office_home} and \ref{tab:ssda_office31_visda}. Our approach outperforms source-free NRC \cite{NRC} and SHOT++ \cite{SHOT++} by 1.5\% and 1.7\% respectively on Office-31 (Table \ref{tab:ssda_office31_visda}), and gives comparable performance to non-source-free works. On the larger and more challenging VisDA dataset, our approach surpasses NRC by 1.6\% and SHOT++ by 1\% (Table \ref{tab:ssda_office31_visda}).  On Office-Home (Table \ref{tab:ssda_office_home}), our model achieves \textit{state-of-the-art} results exceeding the source-free SHOT++ and the non-source-free method SCDA \cite{SCDA} by 1\% and 0.9\% respectively.

\vspace{1mm}
\noindent
\textbf{b) Multi Source Domain Adaptation (MSDA).} 
In Table \ref{tab:msda_compare}, we compare with the source-only baseline (\textit{source-combine}) and source-free works. Even without domain labels, our approach achieves \textit{state-of-the-art} results, even w.r.t.\ non-source-free works on Office-Home (+1\%). On DomainNet, we outperform source-free works (+2.2\%) with comparable results to non-source-free works.

%%%%%%%%%%%%%%%%% Table pretext comparison %%%%%%%%
\begin{wraptable}{r}{0.35\textwidth}
    \centering
    % \vspace{-7mm}
    \setlength{\tabcolsep}{2pt}
    \renewcommand{\arraystretch}{1}
    \vspace{-12mm}
    \caption{Subsidiary task comparisons on \textbf{Office-Home} for source-free DA. Here, baseline is same as \#3 in Table \ref{tab:ablation}.
    }
    \label{tab:subs_task}
    \vspace{1mm}
    \resizebox{0.35\textwidth}{!}{
    \begin{tabular}{lcc}
        \toprule
        Method & SSDA & MSDA \\
        \midrule
        Baseline (B) &  66.2 & 74.3 \\
        B + inpainting & 66.3 & 74.5 \\
        B + colorization &  66.8 & 74.7\\
        \midrule
        B + jigsaw & 67.0  & 74.8\\
        B + patch-loc & 67.6  & 75.0 \\
        B + rotation &  67.9 & 75.4\\
        \midrule
        B + sticker-loc & 68.8  & 75.5 \\
        B + sticker-rot &  69.0 &  75.7\\
        B + sticker-clsf & \textbf{69.7}  & \textbf{76.2} \\
        \bottomrule
    \end{tabular}%
    }
    \vspace{-6mm}
\end{wraptable}
%%%%%%%%%%%%%%%%%%%%%%%%%%%%%%%%%%%%%%%%%%%%%%%%%%%%

\begin{figure*}[t]
    \centering
    \includegraphics[width=\textwidth]{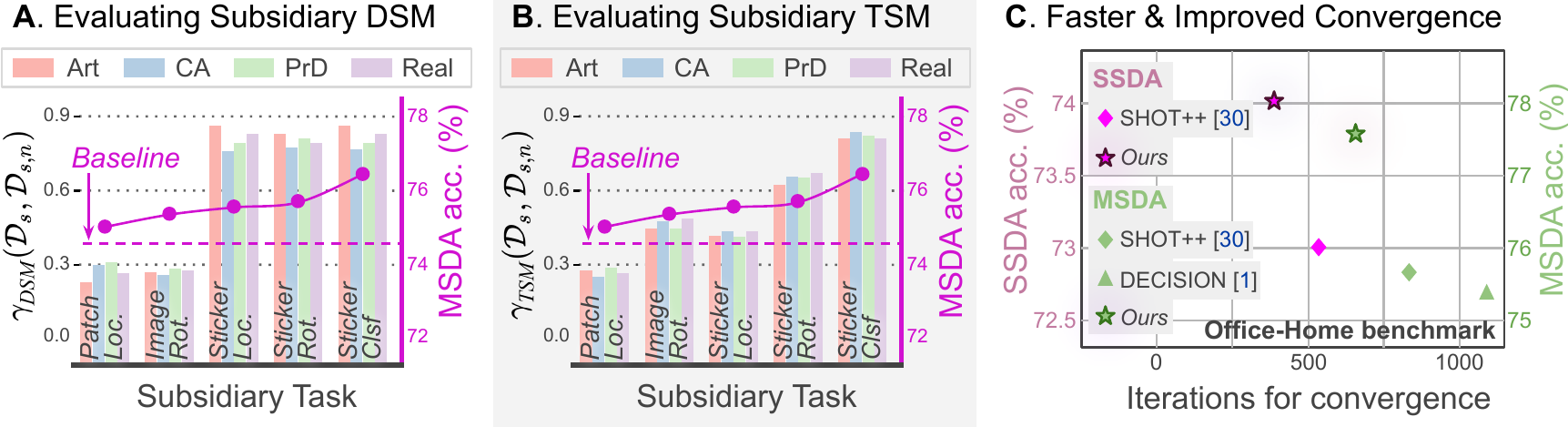}
    \vspace{-7mm}
    \caption{\small
    We observe higher \textbf{A.} domain similarity (DSM) and \textbf{B.} task similarity (TSM) for our sticker-based tasks compared to existing subsidiary tasks like patch-location and image-rotation. This correlates with the better MSDA performance of sticker-based tasks on Office-Home and validates our criteria (Definition \ak{1}).
    \textbf{C.} Faster and improved convergence \wrt prior source-free works on both SSDA and MSDA for Office-Home.
    }
    \vspace{-1mm}
    \label{fig:fig4}
\end{figure*}

\vspace{1mm}
\noindent
\textbf{c) Evaluating the subsidiary DA suitability criteria.}
We empirically evaluate the DSM and TSM for our sticker-based tasks as well as existing tasks borrowed from self-supervised literature in Fig.\ \ref{fig:fig4}\ak{A}, \ref{fig:fig4}\ak{B}. Compared to patch location \cite{sun2019unsupervised} and image rotation \cite{mishra2021surprisingly}, sticker location and sticker rotation tasks exhibit higher DSM and thus, are more suitable with better adaptation performance (also see Table \ref{tab:subs_task}). However, the sticker classification task is the most suitable due to its higher TSM as shape is the primary discriminative features, same as in the goal task. We observe a positive correlation between DA performance and both DSM and TSM, which empirically verifies our suitability criteria. In Table \ref{tab:subs_task}, we additionally compare dense output based tasks like colorization and inpainting, which give marginal gains compared to other subsidiary tasks.

\vspace{1mm}
\noindent
\textbf{d) Faster and improved convergence.}
Fig.\ \ref{fig:fig4}\textcolor{red}{C} illustrates the improved and faster convergence of our approach compared to source-free prior arts for both SSDA and MSDA. The hypothesis space for concurrent subsidiary supervised DA and unsupervised goal task DA, $\mathcal{H}_{g,n}$, is a subset of the hypothesis space for only unsupervised goal task DA, $\mathcal{H}_g^{(uns)}$. Thus, we observe faster convergence for our approach. Further, as per Insight \textcolor{red}{1}, the lower domain discrepancy leads to a lower target error \ie improved convergence.

%%%%%%%%%%%%% Figure Analysis %%%%%%%%%%%%%%%%%%%%%%%%%%
%%%%%%%%%%%%% %%%%%%%%% %%%%%%%%%%%%%%%%%%%%%%%%%%

%%%%%%%%%%%%%%%%%%%%%%%%%%%%% SSDA Office-31, VisDA Table start. %%%%%%%%%%%%%%%%%%%%%%%%%%
\begin{table*}[t]
    \centering
    \setlength{\tabcolsep}{4pt}
    \renewcommand{\arraystretch}{0.95}
    \caption{Single-Source DA (SSDA) on Office-31 and VisDA. SF indicates \textit{source-free}.
    }
    \label{tab:ssda_office31_visda}
    \vspace{-3mm}
    \resizebox{\textwidth}{!}{
    \begin{tabular}{lccccccccccc}
        \toprule
        \multirow{2}{20pt}{\centering Method}& \multirow{2}{*}{\centering SF} 
        &&  \multicolumn{7}{c}{\textbf{Office-31}} && \multicolumn{1}{c}{\textbf{VisDA}} \\
        \cmidrule(lr){4-10}\cmidrule(lr){12-12}
        &&& A$\rightarrow$D & A$\rightarrow$W & D$\rightarrow$W & W$\rightarrow$D & D$\rightarrow$A & W$\rightarrow$A & Avg && S $\rightarrow$ R \\
        \midrule
        CAN~\cite{CAN} & \xmark& & 95.0 & 94.5 & 99.1 & 99.8 & 78.0 & 77.0 & 90.6 & &87.2 \\ % CVPR'19
		FixBi~\cite{FixBi} & \xmark& & 95.0 & 96.1 & 99.3 & 100.0 & 78.7 & 79.4 & 91.4 && 87.2 \\
		CDAN+RADA~\cite{RADA} & \xmark && 96.1 & 96.2 & 99.3 & 100.0 & 77.5 & 77.4 & 91.1 && 76.3\\ %ICCV’21
		RFA~\cite{RFA} & \xmark && 93.0 & 92.8 & 99.1 & 100.0 & 78.0 & 77.7 & 90.2 && 79.4\\ %ICCV’21
		\midrule
		SHOT~\cite{SHOT}& \cmark& & {94.0} &  90.1 &  98.4 & {99.9}  & 74.7 & 74.3 & {88.6} && 82.9\\ % ICML'20
		CPGA~\cite{CPGA} & \cmark && 94.4 & 94.1 & 98.4 & 99.8 & 76.0 & 76.6 & 89.9 && 84.1\\ % IJCAI'21
		HCL~\cite{HCL} & \cmark && 90.8 & 91.3 & 98.2 & 100.0 & 72.7 & 72.7 & 87.6 && 83.5\\ % NeurIPS'21 
		VDM-DA~\cite{VDM-DA}  & \cmark && 93.2 & 94.1 & 98.0 & 100.0 & 75.8 & 77.1 & 89.7 && 85.1\\ % IEEE'21
		${\text{A}^{2}\text{Net}}$~\cite{A2Net} & \cmark && 94.5 & {94.0} & 99.2 & 100.0 & 76.7 & 76.1 & 90.1 && 84.3\\ % ICCV'21
		NRC~\cite{NRC} & \cmark && 96.0 & 90.8 & 99.0 & 100.0 & 75.3 & 75.0 & 89.4 && 85.9\\ % NeurIPS'21
		SHOT++~\cite{SHOT++} & \cmark && 94.3 & 90.4 & 98.7 & 99.9 & 76.2 & 75.8 & 89.2 && {87.3}\\ % TPAMI'21
		\rowcolor{gray!10} \textit{Ours}& \cmark& &
		\textbf{96.1} & \textbf{94.5} & \textbf{99.2} & \textbf{100.0} & \textbf{77.1} & \textbf{78.5} & \textbf{90.9} && \textbf{88.2}\\
        \bottomrule
    \end{tabular}%
    }
    \vspace{-6mm}
\end{table*}

%%%%%%%%%%%%%%%%%%%%%%%%%%%%% SSDA Office-31, VisDA Table ends. %%%%%%%%%%%%%%%%%%%%%%%%%%

\newcolumntype{a}{>{\columncolor{gray!10}}c}

\begin{table}[t]
\begin{minipage}{0.4\textwidth}
    \centering
    \setlength{\tabcolsep}{2.7pt}
    \renewcommand{\arraystretch}{0.95}
    \caption{
    Evaluating compatibility of subsidiary DA with non-source-free DA works on Office-Home. SSDA and MSDA indicate single-source and multi-source DA.
    }
    \label{tab:nonsf_comparison}
    \vspace{-3mm}
    \resizebox{\textwidth}{!}{
    \begin{tabular}{lcc}
        \toprule
        \multirow{2}{*}{Method}
        & \multicolumn{2}{c}{\textbf{Office-Home}} \\
        \cmidrule(lr){2-3} & SSDA & MSDA \\
        \midrule
        CDAN \cite{CDAN} %& \xmark 
        & 65.8 & 69.4\\
        $\;$ + \textit{Subsidiary-DA} %& \cmark 
        & \textbf{67.1} & \textbf{71.2}\\
        \midrule
        SRDC \cite{SRDC} %& \xmark 
        & {71.3} & 73.1\\
        $\;$ + \textit{Subsidiary-DA} %& \cmark 
        & \textbf{71.9} & \textbf{75.2}\\
        \midrule
        FixBi \cite{FixBi} %& \xmark 
        & 72.7 & - \\
        $\;$ + \textit{Subsidiary-DA} %& \cmark 
        & \textbf{73.7} & - \\
        \midrule
        CMSDA \cite{CMSDA}  %& \xmark 
        & - & 76.6 \\
        $\;$ + \textit{Subsidiary-DA} %& \cmark 
        & - & \textbf{78.1} \\
        \bottomrule
    \end{tabular}%
    }

\end{minipage}
\quad
\begin{minipage}{0.57\textwidth}
    \centering
    \setlength{\tabcolsep}{1.4pt}
    \renewcommand{\arraystretch}{1.05}
    \caption{Ablation analysis. Here, \textit{sticker-w-OOS-clsf} denotes learning with all the proposed components unlike in \textit{only-OOS-clsf} (all losses except $\mathcal{L}_{s,n}, \mathcal{L}_{t,n}$) and \textit{only-sticker-clsf} (all losses except $\mathcal{L}_s^{(od)}$). SF denotes source-free constraint. 
    %Ablation study on Office-Home. SF, SSDA and MSDA indicate source-free, single-source DA and multi-source DA. \vspace{10.8pt}
    % \ak{Add source-only baseline and identify comparison rows in slides. Add non-SF setting.}
    }
    \label{tab:ablation}
    \vspace{-3mm}
    \resizebox{\textwidth}{!}{
    \begin{tabular}{lllccccc}
        \toprule
        \multirow{2}{*}{\#} & \multirow{2}{*}{Variation} &&  \multirow{2}{*}{SF} && \multicolumn{3}{c}{\textbf{Office-Home}}\\
        \cmidrule(lr){6-7}
        &&&&& SSDA & MSDA \\
        % \multicolumn{3}{c}{Source-side} & \multicolumn{2}{c}{Target-side} & \multirow{2}{*}{SF} & \multicolumn{2}{c}{\textbf{Office-Home}} \\
        % \cmidrule(lr){2-4} \cmidrule(lr){5-6} \cmidrule(lr){8-9}
        
        % & $\mathcal{L}_{s,g}$ & $\mathcal{L}_{s,n}$ & $\mathcal{L}_{s}^{(od)}$ & $\mathcal{L}_{st}, \mathcal{L}_{div}$ & $\mathcal{L}_{t,n}$ & & SSDA & MSDA \\ 
        \midrule
        % \xmark    & \cmark    & \cmark      & \cmark    & \cmark   & \cmark & \cmark   &  & {71.4} \\
        1. & Source-only baseline && - && 60.2 & 66.9 \\
        2. & \hspace{6pt} + \textit{sticker-w-OOS-clsf} && - && 61.9 & 71.4\\
        \midrule
        3. & Adaptation baseline (B) && \cmark && {66.2} & {74.3}  \\
        4. & B + \textit{only-OOS-clsf} && \cmark && {67.0} & {74.9} \\
        5. & B + \textit{only-sticker-clsf} && \cmark && {{69.7}} & 76.2 \\
        6. & B + \textit{sticker-w-OOS-clsf} && \cmark && 73.1 & {77.6} \\
        7. & B + \textit{sticker-w-OOS-clsf} %(NSF)
        && \xmark && \textbf{74.5} & \textbf{78.3} \\

        \bottomrule
    \end{tabular}%
    }
\end{minipage}
\vspace{-5mm}
\end{table}

% \vspace{1mm}
% \noindent
% \textbf{e) Evaluating subsidiary tasks for concurrent DA.}
% We evaluate existing pretext tasks from self-supervised literature as well as the proposed sticker-based tasks in Table \ref{tab:subs_task}. First, we compare the baseline (\ie performing only source pre-training and goal task adaptation) with dense output based tasks like inpainting and colorization and observe marginal improvements as they have low task-similarity (TSM). Next, we compare

\vspace{-3.5mm}
\subsubsection{Ablation Study.}
\label{subsubsec:ablation}
\vspace{-2mm}
% In Table \ref{tab:ablation}, we report an ablation study on Office-Home to independently analyze the components of our source-side and target-side training.
Below, we discuss a thorough ablation study.
%Table \ref{tab:ablation} reports an ablation study on Office-Home to independently analyze the components of our approach.
% Row (\#1) provide results from the source model, (\#2) from training without the sticker objective, and (\#6) presents our final model. Several conclusions can be drawn from the given table.

\vspace{0.5mm}
\noindent
\textbf{a) Effect of subsidiary supervised DA and
OOS node.}
% First, we assess the target adaption effectiveness of the sticker task supervision. 
In Table \ref{tab:ablation}, we compare the baseline \ie only unsupervised goal task DA (\#3) with the addition of only OOS classifier (\#4). Here, a binary classifier is used for OOS detection. We observe gains of 0.8\% and 0.6\% for SSDA and MSDA respectively. This indicates that only OOS helps, but subsidiary classifier is essential for further improvements.
Next, we compare the baseline (\#3) with concurrent 
% subsidiary supervised DA and goal task DA 
goal-subsidiary DA
without using OOS (\#5). We observe an improvement of 3.5\% and 1.9\% for SSDA and MSDA.
% Using the sticker task objectives in conjunction with the baseline target objectives (\#4) improves the baseline (\#2) performance by 3.5\% and 1.9\%, respectively, for SSDA and MSDA.
% Further adding the OOD node objective to the sticker classifier (\#6) improves the source-target alignment for the backbone features as explained in Sec. \ref{subsec:training_algo}, resulting in performance improvements of 3.1\% and 1.4\% for SSDA and MSDA, respectively.
Adding the OOS objective to the subsidiary supervised DA (\#6 vs. \#4) improves the source-target alignment as explained in Insight \textcolor{red}{6}, resulting in improvements of 3.1\% and 1.4\% for SSDA and MSDA.

\vspace{1.5mm}
\noindent
% \textbf{b) Correlation between subsidiary DA and goal task DA.}
\textbf{b) Subsidiary-goal task similarity.}
% To avoid negative transfer in an unsupervised goal DA task, the subsidiary task should be positively correlated with the goal task.
As per Insight \ak{3}, higher goal-subsidiary task similarity is important for effective learning of both tasks.
Thus, in Table \ref{tab:ablation}, we compare the source-only baseline (\#1) with only subsidiary supervised DA without goal task target adaptation (\#2). We observe gains of 1.7\% and 1.3\% for SSDA and MSDA respectively. This illustrates the positive correlation between sticker classification and goal task even when target goal losses are not used.
% (also for Rw$\shortrightarrow$Pr in Fig.\ \ref{fig:teaser}\textcolor{red}{C}).
% The results for the model developed using only sticker task objectives during target-side training are presented in (\#5) in Table \ref{tab:ablation} which shows an improvement of performance for the source trained model (\#1) by 1.7\% and 1.3\% for SSDA and MSDA, respectively.

% \vspace{1.5mm}
% \noindent
% \textbf{c) Sensitivity to number of sticker classes $\vert \mathcal{C}_n \vert$.} We perform a sensitivity analysis for the number of sticker task categories $\vert \mathcal{C}_n \vert$ for MSDA on Office-Home (Fig.\ \ref{fig:fig4}\textcolor{red}{C}). We observe that the performance improves with increasing number of classes upto 10 and reduces slightly for higher $\vert \mathcal{C}_n \vert$. Overall, we observe consistent gains over the baseline.

% \paragraph{Effects of goal task objectives.}
% The neighbourhood clustering loss seeks to shift each target point in the feature space closer to its neighbour. The model learns tightly clustered features as it moves neighbouring points closer together, resulting in discriminative decision boundaries. When compared to baseline (\#5), combining the clustering objective with sticker task supervision (\#6) considerably improves the model's adaptation capability (by 8.4\% and 5.4\% for SSDA and MSDA, respectively). Additionally, the diversity objective, which was included to avoid trivial solutions (\#7), boosts performance by 2.5\% and 0.8\%, respectively.

% \vspace{1mm}
% \noindent
% \textbf{c) Generalised pretext task.} 
\vspace{-3mm}
\subsubsection{Compatibility with non-source-free DA.}
\vspace{-2mm}
% We provide the results for three non-source free SSDA prior arts~\cite{ganin2016domain,CDAN,SRDC} with our sticker task objectives during training (Table \ref{tab:nonsf_comparison}). 
In Table \ref{tab:nonsf_comparison}, we evaluate the compatibility of concurrent subsidiary supervised DA with existing non-source-free SSDA techniques \cite{ganin2016domain,CDAN,SRDC}.
MSDA results are obtained by combining the multiple sources for each target.
Compared to the original reported results, all four perform better with our proposed subsidiary DA. 
% This demonstrates the general applicability of our approach. Please 
Note that our non-source-free variant outperforms these results ({\#7} in Table \ref{tab:ablation}).

\vspace{-1mm}
\section{Conclusion}

\vspace{-2mm}
In this work, we introduced concurrent subsidiary supervised DA for a pretext-like task to aid the unsupervised goal task DA. We provide theoretical insights to analyze the effect of subsidiary supervised DA on the domain discrepancy and consequently on the goal task adaptation. Based on the insights, we introduce a subsidiary DA suitability criteria to determine DA-assistive subsidiary tasks that improve the goal task DA performance. We also propose a novel sticker intervention based pretext task that follows our criteria. The proposed approach outperforms prior state-of-the-art source-free SSDA and MSDA works on four standard benchmarks, establishing the usefulness of our approach.

\noindent
\textbf{Acknowledgments.}
This work was supported by MeitY (Ministry of Electronics and Information Technology) project (No. 4(16)2019-ITEA), Govt. of India and a research grant by Google.

\clearpage

\appendix

\noindent
\begin{center}
\textbf{\Large Supplementary Material: \\ Concurrent Subsidiary Supervision for Unsupervised Source-Free Domain Adaptation} 
\end{center}

\noindent
\begin{center}
\textbf{\large Supplementary Video} 
\end{center}
We provide a high-level summary video at \url{https://youtu.be/ENJMz-Eg87k}. We visually demonstrate the key insights of our work as well as illustrate the different subsidiary tasks and training algorithm used. We encourage the reader to go through the video for a better understanding of the key ideas.

\noindent
\begin{center}
\textbf{\large Supplementary Document} 
\end{center}

\noindent
In this document, we provide extensive implementation details, additional performance analysis and ablation studies. Towards reproducible research, we release our complete codebase and trained network weights at \url{https://github.com/val-iisc/StickerDA}. This supplementary is organized as follows:

\renewcommand{\labelitemii}{$\circ$}

\begin{itemize}
\setlength{\itemindent}{-0mm}
    \item Section~\ref{sup:sec:notations}: Notations (Table~\ref{sup:tab:notations})
    \item Section~\ref{sup:sec:approach}: Approach (Algo.\ \ref{algo:overall})
    \begin{itemize}
        \setlength{\itemindent}{-0mm}
        \item Target adaptation (Sec.\ \ref{sup:subsec:target_adapt})
        \item Subsidiary DA suitability criteria (Sec.\ \ref{sup:subsec:suitability})
    \end{itemize}
    \item Section~\ref{sup:sec:implementation}: Implementation details
    \begin{itemize}
        \setlength{\itemindent}{-0mm}
        \item Sticker intervention (Sec.\ \ref{sup:subsec:sticker_intervention}, Fig.\ \ref{sup:fig:sticker_intervention}, \ref{sup:fig:oos_data})
        \item Experimental settings (Sec.\ \ref{sup:subsec:experimental})
    \end{itemize}
    \item Section~\ref{sup:sec:analysis}: Analysis
    \begin{itemize}
        \setlength{\itemindent}{-0mm}
        \item Extended comparisons (Sec.\ \ref{sup:subsec:extended_comp}, Table \ref{tab:msda_office-31}, \ref{tab:ssda_office31}, \ref{sup:tab:target_ablation})
        \item Hyperparam.\ sensitivity (Sec.\ \ref{sup:subsec:hyperparam_sensitivity}, Table \ref{sup:tab:sticker_hyperparam_sens}, Fig.\ \ref{fig:sensitivity_num_classes}, \ref{sup:fig:mixup_sens})
        \item Domain discrepancy analysis (Sec.\ \ref{sup:subsec:domain_disc}, Fig.\ \ref{sup:fig:mixup_sens})
        \item Domain alignment analysis (Sec.\ \ref{sup:subsec:domain_align}, Fig.\ \ref{sup:fig:mixup_sens})
        \item Efficiency analysis (Sec.\ \ref{sup:subsec:efficiency}, Table \ref{tab:efficiency})
        \item Combining subsidiary tasks (Sec.\ \ref{sup:subsec:subsidiary_comb}, Table \ref{tab:pretext})
        \item Differences and relationships with prior-arts (Sec.\ \ref{sup:subsec:differences}, Table \ref{tab:characteristics1}, \ref{tab:characteristics2})
    \end{itemize}
\end{itemize}{}

\section{Notations}
\label{sup:sec:notations}
\noindent We summarize the notations used in the paper in Table \ref{sup:tab:notations}. The notations are listed under 5 groups: Models, Preliminaries, Datasets, Samples, and Spaces.

\section{Approach}
\label{sup:sec:approach}

\noindent
We summarize our approach in Algo.\ \ref{algo:overall} and provide details of the target adaptation objectives that were omitted from the main paper due to space constraints.

%%%%%%%%%%%%%%%%%%%% Table notations %%%%%%%%%%%%%%%%%%%%%%
\begin{table}[h]
    \centering
    \caption{\textbf{Notation Table}}
    \label{sup:tab:notations}
    \setlength{\tabcolsep}{12pt}
    \resizebox{0.8\columnwidth}{!}{%
        \begin{tabular}{lcl}
        \toprule
         & \multirow{1}{*}{Symbol} & \multirow{1}{*}{Description} \\
         \midrule
         \multirow{3}{*}{\rotatebox[origin=c]{90}{Models}} & $h$ & Shared backbone feature extractor  \\
         & $f_g$ & Goal task classifier \\
         & $f_n$ & Subsidiary task classifier \\
         \midrule
        \multirow{10}{*}{\rotatebox[origin=c]{90}{Preliminaries}} & $p_s$ & Source marginal distribution \\
         & $p_t$ & Target marginal distribution \\
         & $\epsilon_s$ & Source goal task error\\
         & $\epsilon_t$ & Target goal task error \\
         & $\epsilon_{s,n}$ & Source subsidiary task error\\
         & $\epsilon_{s,n}$ & Target subsidiary task error \\
         & $d_{\mathcal{H}}$ & $\mathcal{H}$-divergence \\
         & $\mathcal{H}$ & Backbone hypothesis space \\
         & $\mathcal{H}_g^{(uns)}$ & $\mathcal{H}$-space for unsup. goal task\\
         & $\mathcal{H}_n^{(sup)}$ & $\mathcal{H}$-space for sup. subsidiary task\\
        \midrule
        \multirow{5}{*}{\rotatebox[origin=c]{90}{Datasets}} & $\mathcal{D}_s$ & Labeled source dataset  \\
         & $\mathcal{D}_t$ & Unlabeled target dataset  \\
         & $\mathcal{D}_{s,n}$ & Subsidiary source dataset  \\
         & $\mathcal{D}_{t,n}$ & Subsidiary target dataset  \\
         & $\mathcal{D}_s^{(od)}$ & Pseudo-OOS dataset  \\
        \midrule
        \multirow{5}{*}{\rotatebox[origin=c]{90}{Samples}} & $(x_s, y_{s})$ & Labeled source sample  \\
        & $(x_{s,n}, y_{s}, y_n)$ & Labeled subsidiary source sample \\
        & $(x_{s}^{(od)}, y_s^{(od)})$ & Labeled pseudo-OOS sample  \\
        & $x_t$ & Unlabeled target sample  \\
        & $(x_{t,n}, y_n)$ & Subsidiary target sample \\
        \midrule 
        \multirow{4}{*}{\rotatebox[origin=c]{90}{Spaces}} 
        & $\mathcal{X}$ & Input space \\
        & $\mathcal{Z}$ & Backbone feature space \\
        & $\mathcal{C}_g$ & Label set for goal task \\
        & $\mathcal{C}_n$ & Label set for subsidiary task \\
        \bottomrule
        \end{tabular}
        }
\end{table}
%%%%%%%%%%%%%%%%%%%% Table ends %%%%%%%%%%%%%%%%%%%%%%%%%%

\subsection{Target adaptation}
\label{sup:subsec:target_adapt}

\noindent
\textbf{Self-training loss.}
We apply self-supervision in the target domain to cluster target samples based on their neighborhood \cite{NRC}. Each target sample in the feature space is aligned with its neighbor. As a result, the model learns a discriminative metric that translates a point to a semantically similar match. This is accomplished by reducing the entropy over point similarity. The model learns tightly clustered features as it moves neighboring points closer together, resulting in discriminative decision boundaries. 

For each mini-batch of target features, we calculate the similarity to all target samples. Let $F_t^{(mb)} \in \mathbb{R}^{|\mathcal{D}_t| \times d}$ denote the memory bank which stores all target features and $d$ denotes the dimensions for output features $f_g \circ h(x_t)$. Here, $|\mathcal{D}_t|$ denotes the number of samples in the target dataset. All stored features are L2-normalized. Specifically,

\begin{equation}
F_t^{(mb)} = [F_1, F_2, \dots , F_{|\mathcal{D}_t|}] 
\end{equation}

where $F_j$ denotes the $j^\text{th}$ item in $F_t^{(mb)}$. 
Let $f_i \!=\! h(x_i)$ denote the features of the current $i^\text{th}$ mini-batch, and $B_t$ denote the set of indices of the mini-batch samples in $F_t^{(mb)}$.
The probability that $f_i$ is a neighbor of the feature $F_j$ is,

\begin{equation}
    p_{i,j} = \frac{\exp(F_j^T f_i / \mathcal{T})}{\sum_{j=1, j\neq i}^{|\mathcal{D}_t|} \exp(F_j^T f_i /\mathcal{T}) }
\end{equation}

where the temperature parameter $\mathcal{T}$ controls the number of neighbors. Then, the entropy \ie the loss is defined as,

\begin{equation}
    \mathcal{L}_{st} = - \frac{1}{|B_t|} \sum_{i\in B_t} \sum_{j=1, j\neq i}^{\mathcal{D}_t} p_{i,j} \log(p_{i,j})
    \label{sup:eqn:loss_st}
\end{equation}

\noindent
\textbf{Diversity loss.}
We encourage the prediction to be balanced to avoid degenerate solutions, where the model predicts all data to a particular class (and does not predict other classes for any target sample). We employ the prediction diversity loss, which has been frequently used in clustering \cite{gomes2010discriminative} and domain adaptation \cite{SHOT}. The diversity objective is,

\begin{align}
    \mathcal{L}_{div}(f_g \circ h(x)) &= D_{KL}(\hat{p}, \frac{1}{|\mathcal{C}_g|} \mathbbm{1}_{|\mathcal{C}_g|}) - \log{|\mathcal{C}_g|}
    \label{sup:eqn:loss_div}
\end{align}

\noindent
where $\mathbbm{1}_{|\mathcal{C}_g|}$ represents a $|\mathcal{C}_g|$-dimensional vector of ones, $\hat{p} \!=\! \expectation_{{x_t} \in \mathcal{D}_t}[\sigma(f_g\!\circ\! h(x_t))]$ is average output embedding for entire target dataset, and $\sigma$ denotes softmax.

\subsection{Subsidiary DA suitability criteria}
\label{sup:subsec:suitability}

\subsubsection{Subsidiary-Domain Similarity Metric (DSM).}
As discussed in Sec.\ \textcolor{red}{3.1.3} of the main paper, we define \textit{subsidiary-domain similarity metric}, $\gamma_\textit{DSM}$ as the inverse of the $\mathcal{H}$-divergence between the two domains. We follow \cite{ganin2016domain} and use the $\mathcal{A}$-distance \cite{ben2006analysis} between the goal task dataset $\mathcal{D}_s$ and the subsidiary task dataset $\mathcal{D}_{s,n}$ as a proxy for $\mathcal{H}$-divergence. We define the dataset labels as 1 for subsidiary source dataset $\mathcal{D}_{s,n}$ and 0 for original source dataset $\mathcal{D}_s$ and train a linear binary classifier on the features of a frozen ImageNet-pretrained \cite{paszke2019pytorch} ResNet-50 \cite{he2016deep_resnet} with a subset of the mixed data, and obtain the classifier error on the other subset as $\psi$. The DSM is then computed as,

\begin{align}
    d_\mathcal{A}(\mathcal{D}_s, \mathcal{D}_{s,n}) =&\; 2\psi (1-\psi)\\
    \gamma_\textit{DSM}(\mathcal{D}_s, \mathcal{D}_{s,n}) =& \; 1\!-\!\frac{1}{2}d_\mathcal{A}(\mathcal{D}_s, \mathcal{D}_{s,n}) 
\end{align}

%%%%%%%%%%%%%%%%%%%% Overall Algorithm %%%%%%%%%%%%%%%%%%%%%%%%%%
\begin{algorithm}[!t]

\caption{Pseudo-code for the proposed approach}
\label{algo:overall}
\begin{algorithmic}[1]

\Statex \underline{\textbf{Source-side training}}

\vspace{1mm}
\State \textbf{Input:} source data $\mathcal{D}_s$, stickered source data $\mathcal{D}_{s,n}$, pseudo-OOS dataset $\mathcal{D}_s^{(od)}$, ImageNet pretrained backbone $h$ (as per \cite{SHOT}), randomly initialized goal classifier $f_g$ and randomly initialized sticker classifier $f_n$.

\vspace{1mm}
\Statex \underline{{\textit{Goal task source pre-training}}}
\vspace{1mm}
\For{$iter < MaxIter$}:
\State Sample batch from $\mathcal{D}_s \cup \mathcal{D}_{s,n}$
\State Compute $\mathcal{L}_{s,g}$ using Eq.\ \textcolor{red}{7} (main paper)
\State \textbf{update} $\theta_h, \theta_{f_g}$ by minimizing $\mathcal{L}_{s,g}$
\EndFor

\vspace{1mm}
\Statex \underline{{\textit{Sticker task source pre-training}}}
\vspace{1mm}
\For{$iter < MaxIter$}:
\State Sample batch from $\mathcal{D}_{s,n}$
\State Sample batch from $\mathcal{D}_s^{(od)}$
\State Compute $\mathcal{L}_{s,n}$ and $\mathcal{L}_s^{(od)}$ using Eq.\ \textcolor{red}{8} (main paper)
\Statex \jnkc{\Comment{using samples from $\mathcal{D}_{s,n}$ and $\mathcal{D}_s^{(od)}$ respectively}}
\State \textbf{update} $\theta_{f_n}$ by minimizing $\mathcal{L}_{s,n}, \mathcal{L}_s^{(od)}$ using separate Adam optimizers
% \Statex \hspace{4.5mm} using separate Adam optimizers
\EndFor

\vspace{1mm}
\Statex \underline{\textbf{Target-side training}}

\vspace{1mm}
\State \textbf{Input:} target data $\mathcal{D}_t$, stickered target data $\mathcal{D}_{t,n}$, source-side pretrained backbone $h$, goal classifier $f_g$ and sticker classifier $f_n$.

\vspace{1mm}
\Statex \underline{{\textit{Source-free target adaptation}}}
\vspace{1mm}
\For{$iter < MaxIter$}:
\State Sample batch from $\mathcal{D}_t$
\State Sample batch from $\mathcal{D}_{t,n}$
\State Compute $\mathcal{L}_{st}$ and $\mathcal{L}_{div}$ using Eq.\ \ref{sup:eqn:loss_st}, \ref{sup:eqn:loss_div} (suppl.)
\Statex \jnkc{\Comment{using samples from both $\mathcal{D}_t$ and $\mathcal{D}_{t, n}$}}
\State Compute $\mathcal{L}_{t, n}$ using Eq.\ \textcolor{red}{9} (main paper)
\Statex \jnkc{\Comment{using samples from only $\mathcal{D}_{t, n}$}}
\State \textbf{update} $\theta_h, \theta_{f_n}$ by minimizing $\mathcal{L}_{t, n}$
\State \textbf{update} $\theta_h$ by minimizing $\mathcal{L}_{st}, \mathcal{L}_{div}$ using separate Adam optimizers
% \Statex \hspace{4.5mm} using separate Adam optimizers
\EndFor

\end{algorithmic}
\end{algorithm}
%%%%%%%%%%%%%%%%%%%% Algorithm ends %%%%%%%%%%%%%%%%%%%%%%%%%%

\noindent
\textbf{How to choose the threshold $\boldsymbol{\zeta_d}$?}
Insight \textcolor{red}{2} introduced a threshold $\zeta_d$ for DSM to select pretext tasks suitable for subsidiary supervised DA. To choose a threshold, we first consider the $\mathcal{A}$-distances between the actual source and target domains. These $\mathcal{A}$-distances are in the range of {1.5 to 2.0} \cite{SImpAl} for Office-Home and indicate the range of $\mathcal{A}$-distances corresponding to realistic domain shifts. This range corresponds to the range of 0 to 0.25 in terms of DSM.
In Fig. \textcolor{red}{7A} of the main paper, we observed DSM in a range of {0 to 0.3} for the patch-location and image-rotation subsidiary task samples \wrt the original samples, indicating that these tasks induce a realistic domain shift. Contrary to this, our proposed sticker task produced DSM in the range of 0.6 to 0.9, indicating much better domain preservation. Thus, we choose the threshold $\zeta_d = 0.5$ which represents $\sim$70\% reduced domain shift \wrt realistic domain shifts (\ie \wrt 1.5 to 2.0).

\subsubsection{Subsidiary-Task Similarity Metric (TSM).}
$\gamma_{\text{TSM}}$ determines how similar a subsidiary task is to the goal task. TSM is calculated using the basic linear evaluation protocol \cite{salman2020do} in self-supervised literature. It illustrates the degree of compatibility between the two tasks. For computing $\gamma_\text{TSM}$, we train a linear classifier $f_n$ on the features $h_{s,g}$ for subsidiary task dataset $\mathcal{D}_{s,n}$ extracted using a frozen source-pretrained ResNet-50 \cite{he2016deep_resnet} backbone. For the sticker classification task, we randomly select 4 classes to keep the number of classes uniform for the different subsidiary task candidates illustrated in Fig.\ \ref{fig:teaser}\textcolor{red}{C} and Fig.\ \ref{fig:fig4}\ak{B} in the main paper. We thus obtain the error for the different subsidiary task classifiers as $\hat{\epsilon}_{s,n}$ and the subsidiary-task similarity metric is computed as:  

\begin{equation}
    \gamma_\textit{TSM}(\mathcal{D}_s, \mathcal{D}_{s,n}) = 1\!-\! \min_{f_n} \hat{\epsilon}_{s,n}(h_{s,g})
\end{equation}

\noindent
\textbf{How to choose the threshold $\boldsymbol{\zeta_n}$?}
Insight \textcolor{red}{3} introduced a threshold $\zeta_n$ for TSM to select pretext tasks suitable for subsidiary supervised DA. The task similarity of the subsidiary task is dependent on the goal task. For computing the threshold for TSM, we plot the $\gamma_\textit{TSM}$ for the candidate subsidiary tasks (Fig.\ \ref{fig:fig4}\ak{B}) and select the appropriate threshold $\zeta_n$. Based on our observations in Fig.\ \ref{fig:fig4}\ak{B} of the main paper, we set $\zeta_n$ as 0.6.

\vspace{0.5mm}
\noindent
\textbf{Suitability criterion.} Definition \textcolor{red}{1} in the main paper gives the overall suitability criterion for selecting the subsidiary task as:
\begin{equation}
    \gamma_\textit{DSM}(\mathcal{D}_s, \mathcal{D}_{s,n})+\gamma_\textit{TSM}(\mathcal{D}_s, \mathcal{D}_{s,n}) > \zeta
    \label{sup:eqn:pretext_task_selection}
\end{equation}
\noindent
Therefore, we set the threshold $\zeta$ as a sum of $\zeta_d$ and $\zeta_n$ \ie 1.1.

%%%%%%%%%%%%% Figure sticker %%%%%%%%%%%%%%%%%%%%%%%%%%
\begin{figure*}[t]
    \centering
    \vspace{-2mm}
    \includegraphics[width=\textwidth]{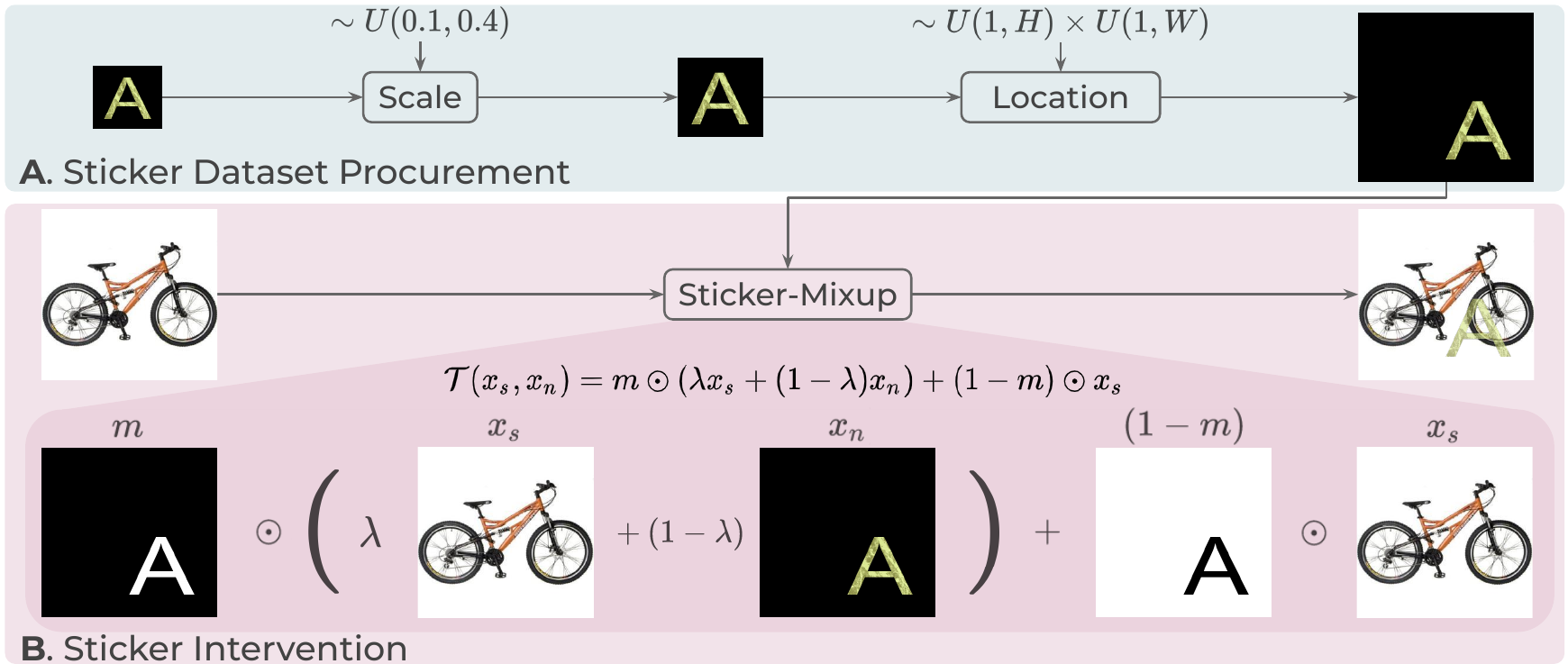}
    \vspace{-6mm}
    \caption{Illustration of \textbf{A.} sticker dataset procurement and \textbf{B.} sticker intervention $\mathcal{T}$ (see Sec.\ \ref{sup:subsec:sticker_intervention}). \textit{Best viewed in color}.
    }
    \vspace{-2mm}
    \label{sup:fig:sticker_intervention}
\end{figure*}
%%%%%%%%%%%%% %%%%%%%%% %%%%%%%%%%%%%%%%%%%%%%%%%%

\section{Implementation details}
\label{sup:sec:implementation}

\subsection{Sticker intervention}
\label{sup:subsec:sticker_intervention}

We define a sticker as a printed alphabet with a random color and random texture \cite{cimpoi14describing} within the alphabet. We scale the sticker randomly and paste it at a random location within a black image (all zeros) with the same size as goal task sample $x_s \!\in\! \mathbb{R}^{H \times W}$, yielding $x_n\! \in \!\mathbb{R}^{H \times W}$ (see Fig.\ \ref{sup:fig:sticker_intervention}\textcolor{red}{A}). The corresponding sticker-task labels $y_n$, along with $x_n$, form the sticker dataset $\mathcal{D}_n$. We also define a pixel-wise mask to perform mixup~\cite{mixup} only at the sticker pixels to avoid  the effects of the black background on the rest of the goal task image.

Specifically, $m(u) \!=\! \mathbbm{1}(x_n(u)\!\neq\!0)$ where $u:[u_x, u_y]$ denotes the spatial index in an $H \times W$ lattice. As shown in Fig.\ \ref{sup:fig:sticker_intervention}\textcolor{red}{B}, a goal task sample $x$, \ie either $x_s$, $x_s^{(od)}$ or $x_t$, and a sticker $x_n$ are combined using mixup \cite{mixup} as,

\begin{equation}
    \mathcal{T}(x, x_n) = m\odot(\lambda x + (1-\lambda)x_n) + (1-m)\odot x
\end{equation}

where $\lambda$ denotes the mixup ratio, $\odot$ represents element-wise multiplication and $\mathcal{T}$ is the sticker intervention (as defined in Insight \ak{4} of main paper).

\subsubsection{Hyperparameters} \hfill \\

\noindent
\textbf{a) Sticker shape} is decided by randomly selected alphabets.

\noindent
\textbf{b) Sticker size} is determined by randomly sampling the size ratio between sticker and goal task images from a uniform distribution over the range $[0.1, 0.4]$. 

\noindent
\textbf{c) Sticker location} for pasting the sticker in the goal task image is sampled from a uniform distribution over the ranges $[1, H]$ and $[1, W]$. The sampled coordinates are rounded down to the nearest integer for pasting the sticker. 

\noindent
\textbf{d) Number of sticker classes} determines the difficulty level of the subsidiary supervised DA problem. 

\noindent
\textbf{e) Mixup ratio} determines the visibility of the sticker \wrt the goal task image. We use a constant mixup ratio of 0.4.

\noindent
We provide ablations for these hyperparameters in Sec.\ \ref{sup:subsec:hyperparam_sensitivity}.

%%%%%%%%%%%%% Figure sensitivity %%%%%%%%%%%%%%%%%%%%%%%%%%
\begin{figure*}[t]
    \centering
    \vspace{-3mm}
    \includegraphics[width=0.8\textwidth]{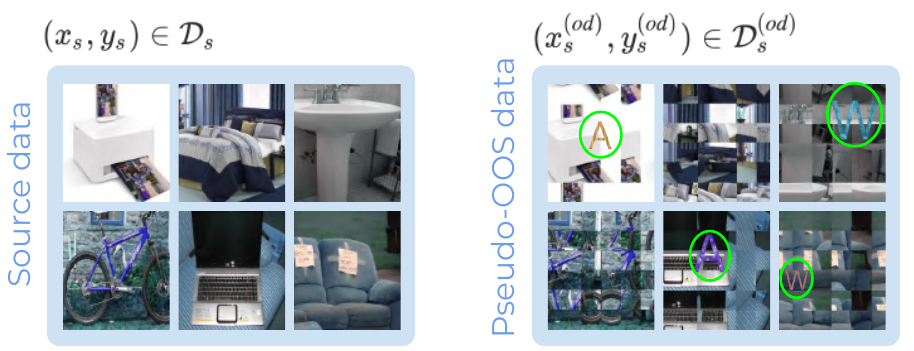}
    \caption{
    The pseudo-OOS data $\mathcal{D}_s^{(od)}$ contains patch-shuffled versions of source data $\mathcal{D}_s$. Green circles only highlight the stickers and are not part of the samples.
    }
    \vspace{-4mm}
    \label{sup:fig:oos_data}
\end{figure*}
%%%%%%%%%%%%% %%%%%%%%% %%%%%%%%%%%%%%%%%%%%%%%%%%

\subsubsection{Usage.}
The intervention is applied in the same manner to both source samples $x_s$ as well as target samples $x_t$, yielding sticker labels $y_n$ for the sticker classifier. Mitsuzumi \etal \cite{mitsuzumi2021generalized} show that, beyond a certain grid size {(4x4)}, shuffling the grid patches makes the domain unrecognizable. Inspired by this, we generate the pseudo-OOS dataset by randomly shuffling the grid patches with a grid size of {(6x6)} as shown in Fig.~\ref{sup:fig:oos_data}. The sticker intervention is also applied to the pseudo-OOS samples in order to emphasize the difference between source and pseudo-OOS samples even when stickers are present. However, for pseudo-OOS samples, the sticker label is treated as $y_s^{(od)}$, for the OOS node to act as an implicit domain discriminator, leading to improved source-target alignment.

\noindent
\textbf{Enabling source-free DA.}
The proposed sticker intervention can be used within source-free constraints. This is because, the alphabet font can be shared between source-side and target-side while the texture dataset \cite{cimpoi14describing} is open-source.

\subsection{Experimental settings}
\label{sup:subsec:experimental}

\noindent
\textbf{Architecture details.} We use a ResNet-50~\cite{he2016deep_resnet} backbone for Office-Home, Office-31 and DomainNet, and ResNet-101 for VisDA, for a fair comparison with prior works. We employ the same network design as SHOT \cite{SHOT}, \ie replacing the classifier with a fully connected layer with batch norm \cite{ioffe2015batch} and another fully connected layer with weight normalization \cite{salimans2016weight}. For the subsidiary classifier, we use the same architecture after ResLayer-3.

\noindent
\textbf{Optimization details.} We employ multiple Adam optimizers during training to avoid loss weighting hyperparameters. Specifically, we use a distinct optimizer for each loss term. In each training iteration, we optimize only one of the losses (round robin method). Each optimizer uses a learning rate of 1e-3. Intuitively, each Adam optimizer's moment parameters adaptively scale the associated gradients, eliminating the requirement for loss-scaling hyperparameter tuning. For source model training, following~\cite{SHOT}, we set the maximum number of epochs to 100 and 30 for Office-31 and Office-Home, whereas it is set to 10 and 15 for VisDA and DomainNet respectively. For adaptation, the maximum number of epochs is set to 15 for all datasets, following \cite{SHOT}.

\section{Analysis}
\label{sup:sec:analysis}

We provide more comparisons with prior state-of-the-art methods and report hyperparameter sensitivity analyses. 

\subsection{Extended comparisons and ablations}
\label{sup:subsec:extended_comp}

\noindent
\textbf{a) Single-Source DA for Office-31 and VisDA.}
Our approach outperforms source-free NRC \cite{NRC} and SHOT++ \cite{SHOT++} by 1.5\% and 1.7\% respectively on Office-31 (Table \ref{tab:ssda_office31}), and gives comparable performance to non-source-free works. On the larger and more challenging VisDA dataset, our approach surpasses NRC by 1.6\% and SHOT++ by 1\% (Table \ref{tab:ssda_office31}).

\noindent
\textbf{b) Multi-Source DA for Office-31.}
To analyze our performance on closed-set MSDA, we compare our approach with source-free and non-source-free prior arts in Table \ref{tab:msda_office-31}. Even without domain labels, our approach achieves \textit{state-of-the-art} results on the Office-31 benchmark, even for the non-source-free setting.

\noindent
\textbf{c) Variance across random seeds.}
We highlight the significance of our results by reporting the mean and standard deviation of accuracy for 5 runs with different random seeds (2nd last row of Table \ref{tab:ssda_office31}) for SSDA. We observe low variance even \wrt prior non-source-free works.

\noindent
\textbf{d) Ablations for target adaptation.}
We present ablations on the goal task objectives for the target-side training ($\mathcal{L}_{st}$ and $\mathcal{L}_{div}$) in Table \ref{sup:tab:target_ablation}. First, we compare the baseline \ie source-trained model (\#1) with the $\mathcal{L}_{div}$ based DA model (\#2). It is interesting to note that only using the diversity objective with subsidiary supervision improves SSDA and MSDA by 2.4\% and 5.5\% respectively over the baseline (\#2 vs. \#1), highlighting the relevance of diversity promotion. 

The neighborhood clustering based self-training loss $\mathcal{L}_{st}$ improves target clustering in the latent $\mathcal{Z}$ space by bringing the backbone features $h(x)$ closer to their respective nearest neighbors. Using $\mathcal{L}_{st}$ in conjunction with the subsidiary DA loss $\mathcal{L}_{t, n}$ enhances the goal task adaptation by 10.5\% and 5.2\% for SSDA and MSDA respectively, compared to not using $\mathcal{L}_{st}$ (\#4 vs. \#2). We observe that employing both $\mathcal{L}_{div}$ and $\mathcal{L}_{st}$ further improves the performance by 3.8\% and 1.9\% for SSDA and MSDA respectively (\#4 vs. \#3), demonstrating that the two losses are complementary for goal task DA.

%%%%%%%%%%%%%%%%%%%%%%%%%%%%% SSDA Office-31 variance Table start. %%%%%%%%%%%%%%%%%%%%%%%%%%
\begin{table*}[t]
    \centering
    \setlength{\tabcolsep}{2pt}
    \caption{Single-Source Domain Adaptation (SSDA) on Office-31 and VisDA benchmarks with mean and standard deviation over 5 runs. The last row indicates the variance over different sets of sticker shapes while others indicate variance over different random seeds. SF indicates \textit{source-free} DA.}
    \vspace{-3mm}
    \label{tab:ssda_office31}
    \resizebox{\linewidth}{!}{
    \begin{tabular}{lccccccccccc}
        \toprule
        \multirow{2}{20pt}{\centering Method}& \multirow{2}{*}{\centering SF} 
        &&  \multicolumn{7}{c}{\textbf{Office-31}} && \multicolumn{1}{c}{\textbf{VisDA}} \\
        \cmidrule(lr){4-10}\cmidrule(lr){12-12}
        &&& A$\rightarrow$D & A$\rightarrow$W & D$\rightarrow$W & W$\rightarrow$D & D$\rightarrow$A & W$\rightarrow$A & Avg && S $\rightarrow$ R \\
        \midrule
		FAA~\cite{huang2021rda} & \xmark && 94.4 & 92.3 & 99.2 & 99.7 & 80.5 & 78.7 & 90.8 && -\\ %ICCV’21
		RFA~\cite{RFA} & \xmark && 93.0 & 92.8 & 99.1 & 100.0 & 78.0 & 77.7 & 90.2 && 79.4 \\ %ICCV’21
		SCDA~\cite{SCDA} & \xmark && 95.4 & 95.3 & 99.0 & 100.0 & 77.2 & 75.9 & 90.5 && -\\ %ICCV’21
		DMRL~\cite{DMRL}& \xmark && 93.4$\pm$0.5  & 90.8$\pm$0.3  & 99.0$\pm$0.2 &100.0$\pm$0.0 & 73.0$\pm$0.3  & 71.2$\pm$0.3  & 87.9 && -\\
		MCC~\cite{MCC}& \xmark&&{98.6}$\pm$0.1 &	{95.5}$\pm$0.2  &98.6$\pm$0.1 &100.0$\pm$0.0 &{72.8}$\pm$0.3  &74.9$\pm$0.3 &{89.4} && - \\
        CAN~\cite{CAN} & \xmark& & 95.0$\pm$0.3 & 94.5$\pm$0.3 & 99.1$\pm$0.2 & 99.8$\pm$0.2 & 78.0$\pm$0.3 & 77.0$\pm$0.3 & 90.6 && 87.2\\
        RWOT~\cite{RWOT}& \xmark&&{94.5}$\pm$0.2  &{95.1}$\pm$0.2  &99.5$\pm$0.2 &100.0$\pm$0.0 &77.5$\pm$0.1  &77.9$\pm$0.3 &90.8 && -\\
		FixBi~\cite{FixBi} & \xmark& & 95.0$\pm$0.4 & 96.1$\pm$ 0.2 & 99.3$\pm$0.2 & 100.0$\pm$0.0 & 78.7$\pm$0.5 & 79.4$\pm$ 0.3 & 91.4 && 87.2\\
		CDAN+RADA~\cite{RADA} & \xmark && 96.1$\pm$0.4 & 96.2$\pm$0.4 & 99.3$\pm$0.1 & 100.0$\pm$0.0 & 77.5$\pm$0.1 & 77.4$\pm$0.3 & 91.1 && 76.3\\ %ICCV’21
		\midrule
		SHOT~\cite{SHOT}& \cmark& & {94.0} &  90.1 &  98.4 & {99.9}  & 74.7 & 74.3 & {88.6} && 82.9\\ % ICML'20
		CPGA~\cite{CPGA} & \cmark && 94.4 & 94.1 & 98.4 & 99.8 & 76.0 & 76.6 & 89.9 && 84.1\\ % IJCAI'21
		HCL~\cite{HCL} & \cmark && 90.8 & 91.3 & 98.2 & 100.0 & 72.7 & 72.7 & 87.6 && 83.5\\ % NeurIPS'21 
		VDM-DA~\cite{VDM-DA}  & \cmark && 93.2 & 94.1 & 98.0 & 100.0 & 75.8 & 77.1 & 89.7 && 85.1\\ % IEEE'21
		${\text{A}^{2}\text{Net}}$~\cite{A2Net} & \cmark && 94.5 & {94.0} & 99.2 & 100.0 & 76.7 & 76.1 & 90.1 && 84.3\\ % ICCV'21
		NRC~\cite{NRC} & \cmark && 96.0 & 90.8 & 99.0 & 100.0 & 75.3 & 75.0 & 89.4 && 85.9\\ % NeurIPS'21
		SHOT++~\cite{SHOT++} & \cmark && 94.3 & 90.4 & 98.7 & 99.9 & 76.2 & 75.8 & 89.2 && {87.3}\\ % TPAMI'21
		3C-GAN~\cite{3C-GAN}& \cmark& & 92.7$\pm$0.4 &  {93.7}$\pm$0.2 &  98.5$\pm$0.1 & 99.8$\pm$0.2  & 75.3$\pm$0.5 & \textbf{77.8}$\pm$0.1 & {89.6} && -\\
		SFDA~\cite{SFDA} & \cmark && 92.2$\pm$0.2 & 91.1$\pm$0.3 & 98.2$\pm$0.3 & 99.5$\pm$0.2 & 71.0$\pm$0.2 & 71.2$\pm$0.2 & 87.2 && - \\ % TAI’21 (Dec)
		\rowcolor{gray!10} \textit{Ours (random seed)}& \cmark& & {95.6}$\pm$0.2 & \textbf{94.6}$\pm$0.2 & \textbf{99.2}$\pm$0.1 & 99.8$\pm$0.2  & {77.0}$\pm$0.3 & 77.7$\pm$0.3 & \textbf{90.7} && \textbf{88.2}$\pm$0.4\\
		\rowcolor{gray!10} \textit{Ours (random sticker)} & \cmark& & {95.5}$\pm$0.1 & 94.2$\pm$0.2 & {98.9}$\pm$0.2 & {99.9}$\pm$ 0.1  & \textbf{77.2}$\pm$0.1 & 76.3$\pm$0.2 & {90.3} && 88.0$\pm$0.3\\
        \bottomrule
    \end{tabular}%
    }
    \vspace{-2mm}
\end{table*}
%%%%%%%%%%%%%%%%%%%%%%%%%%%%% SSDA Office-31 variance Table ends. %%%%%%%%%%%%%%%%%%%%%%%%%%

%%%%%%%%%%%%% Figure sensitivity %%%%%%%%%%%%%%%%%%%%%%%%%%
\begin{figure*}[t]
    \centering
    \includegraphics[width=\textwidth]{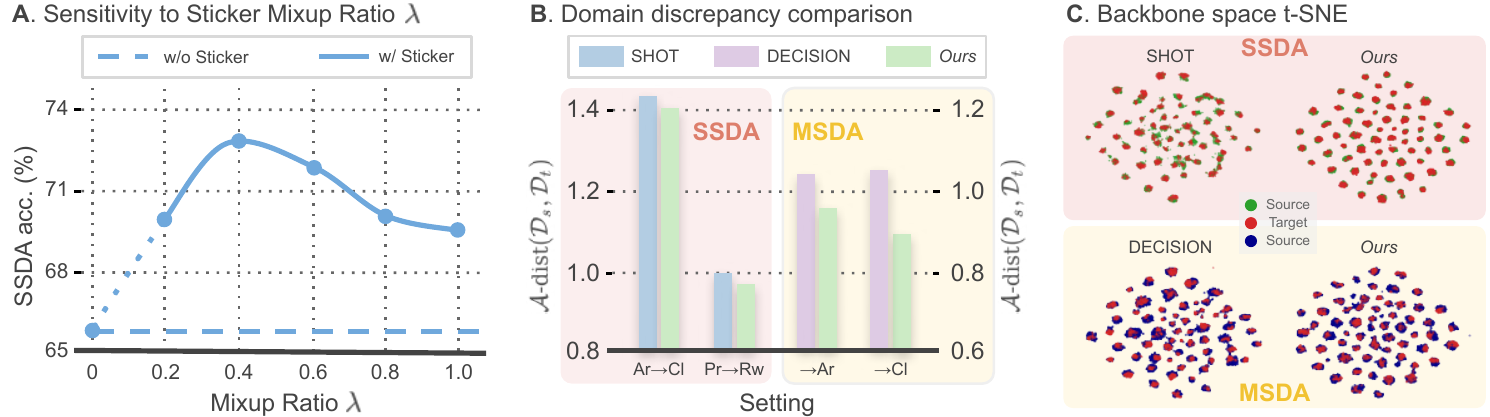}
    \vspace{-6mm}
    \caption{\textbf{A.} Sensitivity to sticker mixup ratio $\lambda$ for SSDA on Office-Home. \textbf{B.} $\mathcal{A}$-distance between source and target data on Office-Home. \textbf{C.} Backbone feature space t-SNE comparisons with SHOT \cite{SHOT} on Rw$\to$Pr (SSDA), DECISION \cite{DECISION} on $\to$Ar (MSDA) from Office-Home.
    }
    \vspace{-4mm}
    \label{sup:fig:mixup_sens}
\end{figure*}
%%%%%%%%%%%%% %%%%%%%%% %%%%%%%%%%%%%%%%%%%%%%%%%%

\subsection{Hyperparameter sensitivity analysis}
\label{sup:subsec:hyperparam_sensitivity}

\noindent
\textbf{a) Sticker shape.} 
We randomly selected 10 alphabets and used them consistently to report all the results in the main paper. However, to test the variance of our approach \wrt sticker shape, we report the mean and standard deviation over 5 runs of SSDA experiments on Office-31 (last row of Table \ref{tab:ssda_office31}), randomly sampling the 10 alphabets (\ie sticker shapes) for each run. We observe a low standard deviation indicating low sensitivity to the sticker shapes.

\noindent
\textbf{b) Sticker size.} We select this scale range based on empirical evidence (Table \ref{sup:tab:sticker_hyperparam_sens}). We observe that adaptation performance suffers with sticker scale less than 0.1, since the sticker is hardly visible, making it difficult for the sticker classifier to receive meaningful supervision. The performance with larger sized stickers (more than 0.7) also drops as the sticker may occlude goal task content significantly.

\noindent
\textbf{c) Sticker location.} We observe that our approach is only mildly sensitive to this hyperparameter (Table \ref{sup:tab:sticker_hyperparam_sens}). We restrict the sticker location to regions far from the image centre and observe slightly lower accuracy. On the other hand, pasting the sticker near the image centre area further decreases performance as the sticker may occlude a larger part of the goal task content. Allowing the sticker to be pasted uniformly across the image yields the best performance.

%%%%%%%%%%%%% Figure Theory %%%%%%%%%%%%%%%%%%%%%%%%%%
\begin{wrapfigure}{r}{0.35\textwidth}
    \centering
    \vspace{-8mm}
    \includegraphics[width=0.35\textwidth]{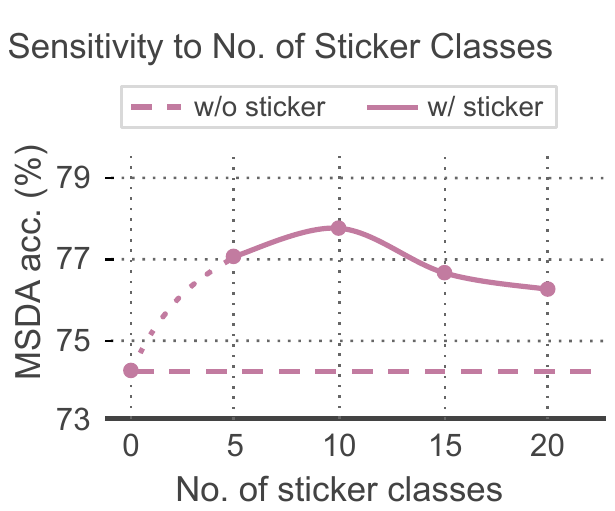}
    \vspace{-7mm}
    \caption{Sensitivity to no.\ of sticker classes $\vert \mathcal{C}_n \vert$ for Office-Home MSDA.}
    \vspace{-6mm}
    \label{fig:sensitivity_num_classes}
\end{wrapfigure}
%%%%%%%%%%%%% %%%%%%%%% %%%%%%%%%%%%%%%%%%%%%%%%%%

\noindent
\textbf{d) Number of sticker classes.} 
We perform a sensitivity analysis for the number of sticker categories $\vert \mathcal{C}_n \vert$ for MSDA on Office-Home (Fig.\ \ref{fig:sensitivity_num_classes}). We observe that performance improves with increasing number of classes upto 10 and reduces slightly for higher $\vert \mathcal{C}_n \vert$. Overall, we observe consistent gains over the baseline.

\noindent
\textbf{e) Mixup ratio $\boldsymbol{\lambda}$.} 
In Fig.\ \ref{sup:fig:mixup_sens}\textcolor{red}{A}, we observe consistent gains over the baseline (mixup ratio $\lambda=0$ \ie sticker classifier and losses not used) for a wide range of $\lambda$ values. The best performance is observed for $\lambda = 0.4$. Intuitively, higher mixup ratios imply very low sticker visibility while lower mixup ratios imply more occlusion of goal task content, both yielding slightly lower performance.

%%%%%%%%%%%%%%%%%%%%%%%%%%%%% Ablation Table start. %%%%%%%%%%%%%%%%%%%%%%%%%%
\begin{table}[t]
\begin{minipage}[b]{0.45\textwidth}
    \centering
    \setlength{\tabcolsep}{5pt}
    \caption{Ablation study on Office-Home. SF, SSDA and MSDA indicate source-free, single-source DA and multi-source DA.
    }
    \vspace{-2mm}
    \label{sup:tab:target_ablation}
    \resizebox{\columnwidth}{!}{
    \begin{tabular}{lcccccc}
        \toprule
        \multirow{2}{*}{\#} & \multicolumn{3}{c}{Target-side} & \multirow{2}{*}{SF} & \multicolumn{2}{c}{\textbf{Office-Home}} \\
        \cmidrule(lr){2-4} \cmidrule(lr){6-7} 
        & $\mathcal{L}_{st}$ & $\mathcal{L}_{div}$ & $\mathcal{L}_{t,n}$ & & SSDA & MSDA \\ 
        \midrule
        1 &  \xmark & \xmark & \xmark & - & {60.2} & 66.9 \\
        2 &  \xmark   & \cmark  & \cmark & \cmark & 62.6 & 72.4 \\
        3 &  \cmark   & \xmark  & \cmark& \cmark & 69.3 & 75.7  \\
        4 & \cmark   & \cmark & \cmark   & \cmark &  \textbf{73.1}  & \textbf{77.6} \\
        \bottomrule
    \end{tabular}%
    }
\end{minipage}
\quad
\begin{minipage}[b]{0.45\textwidth}
    \centering
    \setlength{\tabcolsep}{1.5pt}
    \renewcommand{\arraystretch}{1.2}
    \caption{
    Sensitivity analysis for sticker scale and location on the single-source DA (SSDA) benchmark of Office-Home dataset.
    }
    \vspace{-2mm}
    \label{sup:tab:sticker_hyperparam_sens}
    \resizebox{\textwidth}{!}{
    \begin{tabular}{cccc}
        \begin{tabular}{lc}
            \toprule
            Sticker scale &  Acc.\\
            \midrule
            $0.05 - 0.1$ & 71.8 \\ 
            $0.1 - 0.4$ & 72.2 \\
            $0.4 - 0.7$ & \textbf{73.1} \\ 
            $0.7 - 1.0$ & 72.0 \\
            \bottomrule
        \end{tabular}
    &&&
    \begin{tabular}{lc}
        \toprule
        Sticker location &  Acc.\\
        \midrule
        Central region & 71.5 \\
        Except central region & 72.0 \\ 
        Entire image & \textbf{73.1} \\ 
        \bottomrule
    \end{tabular} 
    \end{tabular}}
\end{minipage}
\end{table}
%%%%%%%%%%%%%%%%%%%%%%%%%%%%% Ablation ends. %%%%%%%%%%%%%%%%%%%%%%%%%%

\subsection{Domain discrepancy analysis}
\label{sup:subsec:domain_disc}

\noindent
In Fig.\ \ref{sup:fig:mixup_sens}\textcolor{red}{B}, we report $\mathcal{A}$-distance as a measure of the domain discrepancy $d_{\mathcal{H}}(p_s, p_t)$ across different source-target pairings in the backbone feature space $\mathcal{Z}$ for our approach and prior source-free state-of-the-art SSDA~\cite{SHOT} and MSDA~\cite{DECISION} works. A lower value for $\mathcal{A}$-distance indicates lower domain discrepancy. In comparison to prior works, our technique clearly achieves lower $\mathcal{A}$-distance between source and target for both settings. This implies that our backbone learns domain-agnostic features that are more generalized to the target domain. This corresponds to an increase in target performance and demonstrates that subsidiary supervised adaptation efficiently minimizes the latent space distribution shift, $d_{\mathcal{H}}(p_s, p_t)$, consistent with Insight \textcolor{red}{1} of the main paper. 

\subsection{Domain alignment analysis}
\label{sup:subsec:domain_align}

\noindent
In Fig.\ \ref{sup:fig:mixup_sens}\textcolor{red}{C}, we present t-SNE \cite{tsne} visualizations of backbone features learned by SHOT \cite{SHOT} and our approach for SSDA, and DECISION \cite{DECISION} and our approach for MSDA. As expected, all three approaches aid the formation of target clusters but source-target alignment for prior arts is weaker compared to our approach. We also observe that our method better preserves the source clusters (\textit{green} in SSDA and \textit{blue} in MSDA) while producing dense clusters for the target features (\textit{red} in both settings) that are better aligned with the source clusters. This improved source-target alignment can be attributed to the OOS node in the sticker classifier, consistent with Insight \textcolor{red}{6} presented in the main paper.

%%%%%%%%%%%%%%%%%%%%%%%%%%%%% Efficiency Table begins. %%%%%%%%%%%%%%%%%%%%%%%%%%

\begin{table}[h]
    \centering
    \setlength{\tabcolsep}{5pt}
    \caption{Training and inference time comparison \wrt NRC~\cite{NRC} and SHOT++~\cite{SHOT++}. All timings are obtained using a single 1080Ti GPU.}
    \vspace{-3mm}
    \label{tab:efficiency}
    \resizebox{\linewidth}{!}{%
        \begin{tabular}{lccccccc}
            \toprule
            \multirow{3}{*}{\begin{tabular}[c]{l}{Method}
            \end{tabular}} & \multicolumn{4}{c}{Training time (in sec), Ar$\to$Cl}  & \multirow{3}{*}{\begin{tabular}{c} Inference \\ time (in \\ millisec) \end{tabular}} & \multicolumn{2}{c}{Office-Home} \\ 
            \cmidrule{2-5} \cmidrule{7-8}
            & \begin{tabular}{c} Source \\ pretrain \end{tabular}  & \begin{tabular}{c} Sticker \\ pretrain \end{tabular} & \begin{tabular}{c} Target \\ adapt \end{tabular} & Total & & \begin{tabular}{c} SSDA \\ Avg. \end{tabular} & \begin{tabular}{c} MSDA \\ Avg. \end{tabular} \\ 
            \midrule
            NRC & 282 & - & {1060} & {1342} & 1.9 & 72.2 & 74.7 \\
            SHOT++ & 306 & - & 10043 & 10349 & 1.9 & 73.0 & 75.7 \\
            \rowcolor{gray!15}\textit{Ours} & 282 & 643 & \textbf{284} & \textbf{1209} & 1.9 & \textbf{74.0} & \textbf{77.6} \\ 
            \bottomrule
        \end{tabular} 
        }
\end{table}
%%%%%%%%%%%%%%%%%%%%%%%%%%%%% Efficiency Table ends. %%%%%%%%%%%%%%%%%%%%%%%%%%

\subsection{Efficiency analysis}
\label{sup:subsec:efficiency}

\noindent
We provide detailed training time comparisons of our work \wrt NRC \cite{NRC} and SHOT++ \cite{SHOT++} in Table \ref{tab:efficiency}. We make certain observations:\ \textbf{1)} We achieve superior target adaptation efficiency with the fastest training (4th column) and the best performance (last 2 columns). Note that we use same learning rate and scheduler as in NRC and SHOT++. \textbf{2)} Inference complexity (6th column) is same for all as we do not require the subsidiary classifier during inference.

% %%%%%%%%%%%%%%%%%%%%%%%%%%%%% Multi-pretext Table starts. %%%%%%%%%%%%%%%%%%%%%%%%%%
% \begin{table}[h]
%     \centering
%     \setlength{\tabcolsep}{14pt}
%     \renewcommand{\arraystretch}{0.75}
%     \caption{
%     Combining multiple subsidiary tasks.}
%     \label{tab:pretext}
%     \resizebox{0.65\linewidth}{!}{
%         \begin{tabular}{lc}
%             \toprule
%             Office-Home & SSDA \\
%             \midrule
%             Baseline (B) &  66.2 \\ \midrule
%             B + patch-loc & 67.6  \\
%             B + rotation &  67.9\\
%             B + rotation + patch-loc & 68.0 \\
%             \midrule
%             B + sticker-rot &  69.0 \\
%             B + sticker-clsf & {69.7} \\
%             B + sticker-rot + sticker-clsf & 69.5 \\
%             \bottomrule
%         \end{tabular}%
%         }
% \end{table}
%%%%%%%%%%%%%%%%%%%%%%%%%%%%% Multi-pretext Table ends. %%%%%%%%%%%%%%%%%%%%%%%%%%

%%%%%%%%%%%%%%%%%%%%%%%%%%%%% MSDA O31 and Combining subs-tasks start. %%%%%%%%%%%%%%%%%%%%%%%%%%
\begin{table}[t]
\begin{minipage}[b]{0.45\textwidth}
    \centering
    \setlength{\tabcolsep}{3pt}
    \caption{ 
    Multi-Source DA (MSDA) comparisons on Office-31.
    }
    \vspace{-3mm}
    \label{tab:msda_office-31}
        \resizebox{\textwidth}{!}{
        \begin{tabular}{lccccc}
            \toprule
            \multirow{2}{*}{Method} & \multirow{2}{*}{SF} & \multicolumn{4}{c}{\textbf{Office-31}} \\
            \cmidrule{3-6}
            & & $\shortrightarrow$A & $\shortrightarrow$W & $\shortrightarrow$D & Avg. \\
            \midrule   
            PFSA~\cite{PFSA} &\xmark  & 57.0 & 97.4 & 99.7 & 84.7 \\
            SImpAl~\cite{SImpAl} &\xmark  & 70.6 & 97.4 & 99.2 & 89.0 \\
            WAMDA~\cite{WAMDA} &\xmark  & 72.0 & 98.6 & 99.6 & 90.0 \\
            MIAN~\cite{MIAN} & \xmark & 76.2 & 98.4 & 99.2 & 91.3 \\ % ICCV'21 
            MLAN~\cite{MLAN} & \xmark & 75.7 & 98.8 & 99.6 & 91.4 \\ % WACV'21
            \midrule
            {Source-combine} &\xmark  & 65.2 & 94.6 & 98.4 & 86.1\\
            SHOT\cite{SHOT}-Ens  &\cmark &{75.0} &{94.9} &{97.8} & {89.3}\\
            DECISION~\cite{DECISION} &\cmark & {75.4} &{98.4} &{99.6} & {91.1}\\
            CAiDA~\cite{caida} &\cmark & {75.8} &{98.9} &{99.8} & {91.6}\\
            \rowcolor{gray!10}Ours &\cmark &\textbf{78.3} &\textbf{99.1} &\textbf{99.7} &\textbf{92.4}\\
            \bottomrule
        \end{tabular} 
        }
\end{minipage}
\quad
\begin{minipage}[b]{0.45\textwidth}
    \centering
    \setlength{\tabcolsep}{8pt}
    \renewcommand{\arraystretch}{0.75}
    \caption{
    Combining multiple subsidiary tasks (SSDA on Office-Home).}
    \vspace{-3mm}
    \label{tab:pretext}
    \resizebox{\linewidth}{!}{
        \begin{tabular}{lc}
            \toprule
             & SSDA \\
            \midrule
            Baseline (B) &  66.2 \\ \midrule
            B + patch-loc & 67.6  \\
            B + rotation &  67.9\\
            B + rotation + patch-loc & 68.0 \\
            \midrule
            B + sticker-rot &  69.0 \\
            B + sticker-clsf & {69.7} \\
            B + sticker-rot + sticker-clsf & 69.5 \\
            \bottomrule
        \end{tabular}%
        }
\end{minipage}
\vspace{-4mm}
\end{table}
%%%%%%%%%%%%%%%%%%%%%%%%%%%%% MSDA O31 and Combining subs-tasks ends. %%%%%%%%%%%%%%%%%%%%%%%%%%

\noindent
\subsection{Combining subsidiary tasks}
\label{sup:subsec:subsidiary_comb}

Introducing multiple subsidiary tasks in the same framework brings up additional challenges like multi-task balancing. For instance, consider a combination of rotation (\texttt{Rot}) and patch-location (\texttt{PL}). From Fig.\ \ref{fig:teaser}\textcolor{red}{C} in the main paper, \texttt{Rot} has high TSM while \texttt{PL} has high DSM. This does not imply that combining \texttt{Rot} and \texttt{PL} would yield a better overall TSM+DSM, and may rather have a detrimental impact. Thus, one should aim for a subsidiary task having both TSM and DSM greater than those of \texttt{Rot} and \texttt{PL}. Empirically, we do not find any conclusive result. In Table \ref{tab:pretext}, we observe that while \texttt{Rot}+\texttt{PL} shows marginal gains, combining Sticker-rot and Sticker-clsf shows degraded performance.

%%%%%%%%%%%%%%%%%%%%%%%%%%%% characteristics table1 %%%%%%%%%%%%%%%%%%%%%%%%%%
\begin{table}[h]
    \centering
    \renewcommand{\arraystretch}{0.95}
    \setlength{\tabcolsep}{12pt}
    \caption{{Comparisons \wrt pretext task based DA works.}
    }
    \vspace{-3mm}
    \label{tab:characteristics1}
    \resizebox{\columnwidth}{!}{
    \begin{tabular}{lccc}
        \toprule
        Method & \begin{tabular}{c} Pretext Task \end{tabular} &  \begin{tabular}{c} High \\ DSM+TSM  \end{tabular} & \begin{tabular}{c} Additional \\ regularization
        \end{tabular} \\
        \midrule
        
        \rowcolor{gray!20} & Rotation, & &   \\ 
        \rowcolor{gray!20}\multirow{-2}{*}{SS-DA [\textcolor{darkpastelgreen}{17}]} 
        & Rot. Patch Jigsaw & \multirow{-2}{*}{\xmark} & \multirow{-2}{*}{\begin{tabular}{c} Adv. alignment, \\ AdaBN\end{tabular}} \\
        
        JiGen [\textcolor{darkpastelgreen}{5}]  & Jigsaw & \xmark & Augmentations \\

        \rowcolor{gray!20}PAC [\textcolor{darkpastelgreen}{34}] & Rotation & \xmark & Aug. consistency \\
        
         \textit{\textbf{Ours}} & \textbf{Sticker} & \cmark & \textbf{None}   \\
        \bottomrule
    \end{tabular}%
    }
\end{table}
%%%%%%%%%%%%%%%%%%%%%%%%%%%%% characteristics table1 %%%%%%%%%%%%%%%%%%%%%%%%%%

%%%%%%%%%%%%%%%%%%%%%%%%%%%% characteristics table2 %%%%%%%%%%%%%%%%%%%%%%%%%%
\begin{table}[h]
    \centering
    \renewcommand{\arraystretch}{1.1}
    \setlength{\tabcolsep}{10pt}
    \caption{{Comparisons \wrt prior source-free DA works.}
    }
    \vspace{-3mm}
    \label{tab:characteristics2}
    \resizebox{\columnwidth}{!}{
    \begin{tabular}{llc}
        \toprule
        Method & \begin{tabular}{c} Key insights (differences) \end{tabular} &  \begin{tabular}{c} Common \end{tabular} \\
        \midrule
        \rowcolor{gray!20} {\footnotesize SHOT} & Info-max.\ for implicit feature alignment & {\footnotesize$\mathcal{L}_{div}$} \\
        {\footnotesize SHOT++} & Easy-hard target split for better adaptation & {\footnotesize$\mathcal{L}_{div}$} \\
        \rowcolor{gray!20} {\footnotesize CPGA} & Contrastive prototypes for better pseudo-labels & {\footnotesize$\mathcal{L}_{st}$} \\
        {\footnotesize GSFDA} & Local struct.\ clustering for better repr.\ learning & {\footnotesize$\mathcal{L}_{st}, \mathcal{L}_{div}$} \\
        \rowcolor{gray!20} {\footnotesize NRC} & Cluster assumption for better pseudo-labels & {\footnotesize$\mathcal{L}_{st}, \mathcal{L}_{div}$} \\
        \multirow{2}{*}{\footnotesize A$^2$Net} & Dual classifiers to find src-similar tgt samples and & 
        \multirow{2}{*}{-} \\
        \addlinespace[-0.5ex] % to reduce space between two rows
         & contrastive matching for category-wise alignm.\ &  \\
        \rowcolor{gray!20} & \textbf{1.}\ How and when subsidiary task is DA-assistive? &  \\
        \rowcolor{gray!20} & \textbf{2.}\ Criteria for DA-assistive subsidiary tasks &  \\
         \rowcolor{gray!20}\multirow{-3}{*}{\textit{\textbf{Ours}}} & \textbf{3.}\ Process of sticker intervention & \multirow{-3}{*}{\footnotesize$\mathcal{L}_{st}, \mathcal{L}_{div}$} \\
        \bottomrule
    \end{tabular}%
    }
\end{table}
%%%%%%%%%%%%%%%%%%%%%%%%%%%%% characteristics table2 %%%%%%%%%%%%%%%%%%%%%%%%%%

\noindent
\subsection{Differences and relationships with prior-arts} 
\label{sup:subsec:differences}

These are discussed in Table \ref{tab:characteristics1} and \ref{tab:characteristics2}. Our method is free from additional regularization unlike prior works (Table \ref{tab:characteristics1}). While our key contributions are unique, the common losses are widely used (\eg\ GSFDA, NRC in Table \ref{tab:characteristics2}).

{
% \small
\bibliographystyle{splncs04}
\bibliography{egbib}
}
\end{document}